\documentclass[10pt,twocolumn,letterpaper]{article}

\usepackage{wacv}              

\usepackage{url}
\usepackage{booktabs}
\usepackage{amsfonts}
\usepackage{nicefrac}
\usepackage{microtype}
\usepackage{xcolor}
\usepackage{lipsum}

\definecolor{dblue}{RGB}{52, 104, 192}
\definecolor{dgreen}{RGB}{65, 171, 93}
\definecolor{dred}{RGB}{210, 69, 69}
\definecolor{dpurple}{RGB}{148, 103, 189}
\definecolor{dorange}{RGB}{255, 127, 14}
\definecolor{dcyan}{RGB}{35, 70, 100}

\definecolor{color_cem}{RGB}{102, 194, 165}
\definecolor{color_ecbm}{RGB}{252, 141, 98}
\definecolor{color_ar}{RGB}{141, 160, 203}
\definecolor{color_scbm}{RGB}{231, 138, 195}

\usepackage{bbm}
\usepackage{graphicx}
\usepackage{amsmath}
\usepackage{amssymb}
\usepackage{amsthm}
\usepackage{multirow}
\usepackage{caption}
\usepackage{wrapfig}
\usepackage{enumitem}
\definecolor{wacvblue}{rgb}{0.21,0.49,0.74}
\usepackage[pagebackref,breaklinks,colorlinks,allcolors=wacvblue]{hyperref}
\usepackage{subcaption}
\usepackage{multicol}
\usepackage{colortbl}
\usepackage{xspace}
\usepackage{siunitx}
\usepackage{array}
\usepackage{pifont}
\usepackage{algorithm}
\usepackage{algpseudocode}

\newcommand{\cmark}{\textcolor{dblue}{\ding{51}}}%
\newcommand{\xmark}{\textcolor{dred}{\ding{55}}}%

\definecolor{tabgray}{HTML}{f2f2f2}
\definecolor{tablavender}{HTML}{E6F7E6}
\definecolor{tabnavy}{HTML}{00008B}
\definecolor{tabwhite}{HTML}{FFFFFF}
\definecolor{linf}{HTML}{8172b2}
\definecolor{l2}{HTML}{ccb974}
\definecolor{l1}{HTML}{64b5cd}

\newcommand{\cando}{\textsc{CanDO}\xspace}

\definecolor{lavender}{HTML}{9ad756}
\definecolor{DBlue}{RGB}{31, 119, 180}
\definecolor{DGreen}{RGB}{44,160,44}
\definecolor{DRed}{RGB}{214,39,40}
\definecolor{dorange}{RGB}{255, 127, 14}
\definecolor{dorange2}{RGB}{255, 165, 0}
\usepackage{bm, dsfont}

\let\hat\widehat

\newcommand{\be}{\bm{e}}

\newcommand{\bg}{\bm{g}}

\newcommand{\bp}{\bm{p}}

\newcommand{\Cb}{\mathbf{C}}

\newcommand{\Ib}{\mathbf{I}}

\newcommand{\Pb}{\mathbf{P}}

\newcommand{\Rb}{\mathbf{R}}

\newcommand{\Tb}{\mathbf{T}}

\newcommand{\Vb}{\mathbf{V}}

\newcommand{\cA}{\mathcal{A}}

\newcommand{\cE}{\mathcal{E}}

\newcommand{\bepsilon}{\bm{\epsilon}}

\newcommand{\btheta}{\bm{\theta}}

\newcommand{\bmu}{\bm{\mu}}

\newcommand{\bDelta}{\bm{\Delta}}

\newcommand{\argmin}{\mathop{\mathrm{argmin}}}
\newcommand{\argmax}{\mathop{\mathrm{argmax}}}

\makeatletter
\newcommand{\ve}{\@ifnextchar\bgroup{\velong}{{\bm{e}}}}
\newcommand{\velong}[1]{{\bm{#1}}}
\makeatother

\def\wacvPaperID{776} 
\def\confName{WACV}
\def\confYear{2027}

\title{Recovering Black-Box Point Cloud Models from Pose Shifts\\via Derivative-Free Optimization}

\author{Seonghwan Park, Min-Gyu Park\\
Korea Electronics Technology Institute (KETI)\\
{\tt\normalsize \{shpark97, mpark\}@keti.re.kr}
\vspace{0.5em}
}

\begin{document}
\maketitle
\begin{abstract}
Point cloud models are increasingly deployed through score-only ``black-box'' interfaces.
Under this access model, test-time rotations outside the training distribution can substantially degrade performance.
Although existing mitigations such as equivariant retraining and gradient-based adaptation are effective, they are infeasible without access to model weights or gradients.
To address this limitation, we propose \cando, a test-time canonicalization method that recovers lost performance using only forward queries.
Specifically, \cando searches for an input transformation that maximizes model confidence and uses derivative-free optimization to perform this search without gradients.
We further identify three conditions for successful recovery: a training augmentation gap, an accuracy-aligned confidence surrogate, and global search over a multi-modal objective.
Across multiple architectures and benchmarks, \cando recovers $76\%$ of the oracle gap under azimuthal misorientation and $51\%$ under arbitrary $SO(3)$ rotations for object part segmentation.
These results show that frozen score-only point cloud models can recover from rotation-induced distribution shifts through post-deployment canonicalization alone.
\end{abstract}
\addtocontents{toc}{\protect\setcounter{tocdepth}{0}}
\section{Introduction}
\label{sec:introduction}
Point clouds are the native data representation for depth and LiDAR sensing in autonomous driving, robotics, and augmented reality~\citep{geiger2012kitti,caesar2020nuscenes}. 
Deep neural networks have become the de facto standard for processing them, from the PointNet family~\citep{qi2017pointnet,qi2017pointnetpp} to modern convolutional and transformer backbones~\citep{thomas2019kpconv,zhao2021pointtransformer,qian2022pointnext,guo2021survey}. 
As these models move into large-scale deployment, they are increasingly exposed through score-only \emph{black-box} interfaces that return probabilities or logits but provide no access to weights or gradients~\citep{tramer2016stealing}.

A key challenge under this access model is rotation-induced distribution shift.
Point cloud models are typically trained on canonical poses or limited orientation augmentations, biasing representations toward the training orientation distribution~\citep{esteves2018,deng2021vn,zhang2019riconv}.
At test time, rotations alter the coordinate representation while preserving semantic content, creating a train-test orientation mismatch.
For standard backbones without built-in rotation invariance or equivariance, this mismatch can substantially degrade predictions~\citep{esteves2018,deng2021vn}.

Addressing this rotation-induced shift requires capabilities unavailable after black-box deployment: rotation-invariant or equivariant architectures and learned canonicalizers require pre- or auxiliary training~\citep{zhang2019riconv,deng2021vn,kaba}, test-time adaptation requires gradients or mutable internal statistics~\citep{tent,memo,sun2020ttt}, and generative-prior methods rely on large external models~\citep{cloudfixer,focal}.
These requirements are incompatible with frozen score-only models, where adaptation relies solely on forward queries.
This motivates our central question:
\emph{``Can a frozen point cloud model recover from rotation-induced distribution shift using only forward queries?''}

\begin{figure*}[t]\centering
\begin{subfigure}{0.475\linewidth}\centering
\includegraphics[width=\linewidth]{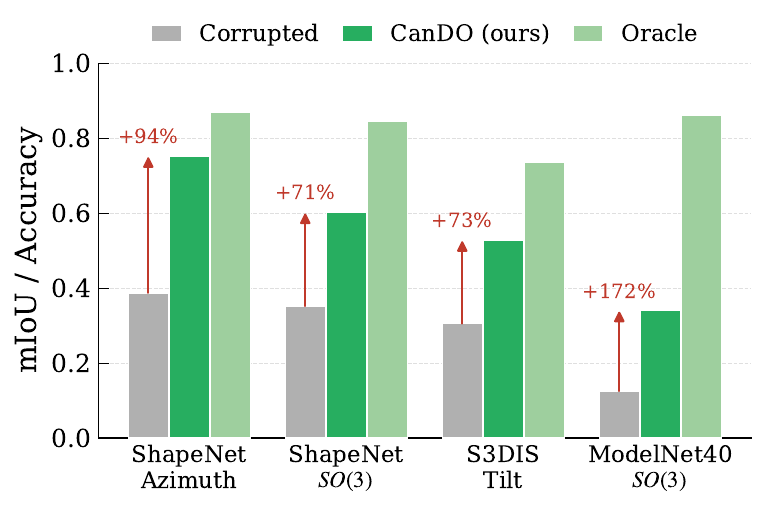}
\caption{Recovery from rotation-induced shifts}
\label{fig:pitch_a}
\end{subfigure}
\begin{subfigure}{0.475\linewidth}\centering
\includegraphics[width=\linewidth]{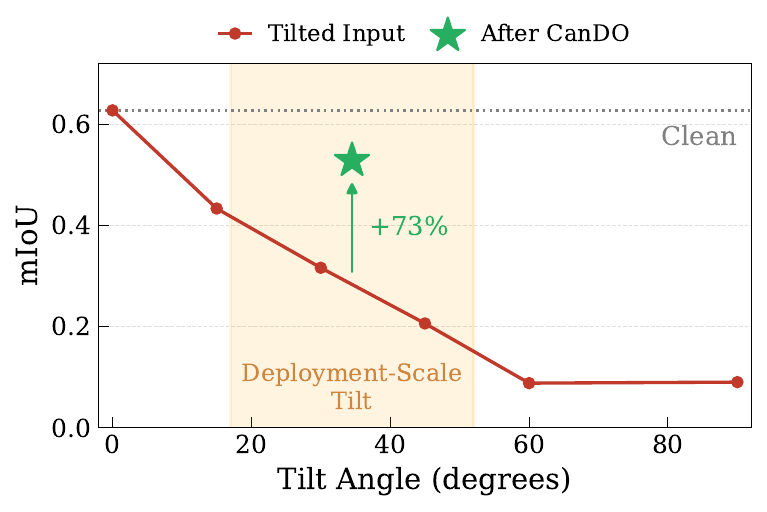}
\caption{Augmentation gap under tilt}
\label{fig:pitch_b}
\end{subfigure}
\vspace{-0.5em}
\caption{
    (a) Rotation-induced shifts within each frozen point cloud model's augmentation gap collapse performance, and \cando recovers a large share of the gap to the supervised oracle from forward queries alone.
    (b) A frozen point cloud model on S3DIS collapses under tilt despite yaw augmentation; the star marks the mIoU recovered by \cando over the shaded deployment band, without labels or gradients.
    }
\label{fig:pitch}
\end{figure*}

Answering this inquiry necessitates optimization without gradients. 
Since the black-box restricts internal access, we turn to derivative-free optimization (DFO)~\citep{kolda2003optimization, conn2009introduction, rios2013derivative}, which operates exclusively on the scalar outputs of evaluated samples. 
However, its query efficiency deteriorates in high-dimensional spaces~\citep{ghadimi2013stochastic,spall1992spsa,duchi2015optimal}, making direct optimization over point coordinates or model parameters impractical.
A viable black-box adaptation method must therefore exploit a low-dimensional structure in the test-time shift.

Rotation-induced shifts naturally provide such a structure.
Rotations in $SO(3)$ have three degrees of freedom and can be represented by standard parameterizations, \eg, Rodrigues vectors~\citep{rodrigues1840,murray1994mathematical} or constrained unit quaternions~\citep{shoemake1985}.
We exploit this compact search space with \cando, a test-time \underline{Can}onicalization method via \underline{D}F\underline{O}.
Given a frozen score-only point cloud model, \cando searches over candidate rotations and selects the transformation that maximizes the model's confidence as a label-free surrogate objective.
A do-no-harm identity prior preserves the original orientation unless another transformation yields a higher score.
Thus, \cando performs post-deployment canonicalization using only forward queries, without labels, gradients, or weights, to mitigate rotation-induced distribution shifts (see \Cref{fig:pitch}).

Beyond the method itself, we analyze when such recovery is possible.
We identify three key factors: (i) an augmentation gap, where the model remains sensitive to rotations absent from training; (ii) confidence-accuracy alignment, where confidence serves as a proxy for task performance; and (iii) global search, which avoids poor solutions in non-convex landscapes.
These analyses also clarify when confidence-only adaptation suffices and when additional task-specific constraints or external priors are required.

Our key contributions are threefold:
\begin{itemize}[leftmargin=1em, topsep=-\parskip, itemsep=0.1em]

    \item We formalize the score-only black-box adaptation setting, where a frozen 3D point cloud model must overcome severe pose shifts relying strictly on forward queries.

    \item We propose \cando, a test-time canonicalization method that leverages DFO to maximize model confidence, recovering lost performance without labels or gradients.

    \item We demonstrate that \cando outperforms prior forward-only methods across multiple tasks, scales, and architectures. 
        We complement this empirical success with a rigorous analysis of the conditions required for recovery and the limits of a self-contained optimization objective.

\end{itemize}
\section{Related Work}
\label{sec:related_work}

\paragraph{Point cloud models and robustness.}
Point cloud models have evolved through diverse architectures, including point-based networks~\citep{qi2017pointnet,qi2017pointnetpp,qian2022pointnext,deng2022pointvector}, graph and convolutional models~\citep{wang2019dgcnn,thomas2019kpconv}, transformer backbones~\citep{zhao2021pointtransformer,pang2022pointmae}, and rotation-equivariant architectures~\citep{deng2021vn}.
Despite their strong performance, these models remain vulnerable to corruptions, acquisition noise, and distribution shifts~\citep{pointcloudc,uy2019scanobjectnn}.
Among these, rotations are particularly challenging because they preserve semantic labels while changing the coordinate frame, exposing a mismatch between the learned orientation bias and the test-time orientation distribution~\citep{esteves2018,deng2021vn}.

\paragraph{Rotation robustness.}
Rotation robustness is commonly addressed through architectural design or canonicalization.
Invariant~\citep{zhang2019riconv} and equivariant architectures~\citep{esteves2018,thomas2018tfn,deng2021vn} encode rotation robustness into the network, thus requiring pre-deployment model design and training.
Canonicalization maps inputs to a reference orientation, either with parameter-free heuristics prone to sign, symmetry, and degeneracy ambiguities~\citep{xiao2020pca}, or with learned modules trained to predict canonical frames~\citep{kaba,sun2021canonical,sajnani2022condor,spezialetti2020compass}.
These methods do not directly apply to deployed models without architecture changes, retraining, or an additional trained component.

\paragraph{Test-time adaptation.}
Test-time adaptation (TTA) adapts models on unlabeled test data by updating parameters, normalization statistics, or auxiliary modules.
Common objectives include self-supervision~\citep{sun2020ttt}, entropy minimization~\citep{tent,memo}, information maximization~\citep{liang2020shot}, and normalization updates~\citep{mirza2022dua,mate}.
Backpropagation-free TTA reduces computation but still assumes parameter access or trainable components~\citep{wang2024bftt3d}.
Input-space methods keep the task model fixed, but CloudFixer~\citep{cloudfixer} and FOCAL~\citep{focal} rely on external diffusion or foundation-model priors.
In contrast, \cando uses only the deployed model's score outputs.

\paragraph{Derivative-free optimization.}
Derivative-free optimization (DFO) solves optimization problems using only queried function values.
Representative families include evolutionary strategies~\citep{wierstra2014nes,hansen2001cmaes,salimans2017es}, Bayesian optimization~\citep{shahriari2015taking}, and finite-difference or simultaneous-perturbation methods~\citep{spall1992spsa,ghadimi2013stochastic,duchi2015optimal,liu2020zoo}.
DFO has been used for black-box adversarial attacks~\citep{chen2017zoo,ilyas2018blackbox} and black-box tuning of pretrained models~\citep{sun2022bbt,sun2022bbtv2,park2025zip}; we apply it to post-deployment point cloud canonicalization over low-dimensional rotation parameters.
\section{Proposed Method}
\label{sec:method}

In this section, we present \cando, a test-time canonicalization method that restores the performance of frozen point cloud black-box models using only forward queries.
We first formalize the black-box problem setting (\Cref{subsec:formulation}) and then parameterize rotation-induced shifts as low-dimensional input transformations (\Cref{subsec:transform}).
Without labels or gradients, we score candidate transformations with a confidence surrogate and a do-no-harm prior (\Cref{subsec:objective}).
We optimize this objective using an evolution strategy (\Cref{subsec:cma}).

\subsection{Problem Formulation}
\label{subsec:formulation}

Let $g$ be a frozen point cloud model exposed only through score-based forward queries.
Each query returns class probabilities or logits.
Given an input point cloud $\Pb=\{\bp_i\}_{i=1}^{N}$ with $N$ points and $\bp_i\in\mathbb{R}^3$, our goal is to prepend an input transformation $\Tb$ that maximizes downstream task performance.
If the task loss $\ell$ and ground-truth target $y$ were available, the ideal per-input objective would be
\begin{equation}
  \Tb^*(\Pb) \;=\; \argmin_{\Tb}\;
  \ell\!\left(g(\Tb(\Pb)),\,y\right),
  \label{eq:ideal}
\end{equation}
with $g$ held fixed. 
However, the black-box regime imposes two critical constraints that prevent standard optimization. 
First, because we lack access to the model's internal weights, no gradient of $\ell$ can propagate through $g$; thus, $\mathbf{T}$ cannot be optimized via backpropagation. 
Second, labels are unavailable at test-time, rendering the evaluation of $\ell$ impossible. 

\subsection{Parameterizing the Input Transform}
\label{subsec:transform}

The formulation above does not specify the transformation $\Tb$.
Motivated by the low-dimensional structure of rotations, we instantiate $\Tb$ as a global rotation of the input point cloud.
Although $SO(3)$ admits standard parameterizations including unit quaternions~\citep{shoemake1985} and axis-angle vectors~\citep{rodrigues1840,murray1994mathematical}, we use the Rodrigues form as a direct three-parameter representation.
This matches the intrinsic degrees of freedom of $SO(3)$ while avoiding constrained quaternion optimization.
Let $\btheta\in\mathbb{R}^3$ be an axis-angle vector with rotation angle $\phi=\|\btheta\|$.
We obtain the rotation matrix $\Rb_{\btheta}\in SO(3)$ using Rodrigues' formula:
\begin{equation}
  \Rb_{\btheta}
  =
  \Ib
  + \frac{\sin \phi}{\phi}[\btheta]_\times
  + \frac{1-\cos \phi}{\phi^2}[\btheta]_\times^2,
  \label{eq:rodrigues}
\end{equation}
where $[\cdot]_\times$ denotes the skew-symmetric cross-product matrix.
We use the standard limit at $\phi\!=\!0$.
For a point cloud represented as row vectors, the transformed input is
\begin{equation}
  \Tb_{\btheta}(\Pb) \;=\; \Pb \Rb_{\btheta}^{\top}.
  \label{eq:transform}
\end{equation}

This formulation covers all transformation families considered in our experiments.
Restricting $\btheta$ to the gravity axis yields a one-dimensional azimuthal search, restricting the axis to the horizontal plane yields a two-dimensional tilt search, and leaving $\btheta$ unconstrained gives full $SO(3)$ search.
For isotropic scaling, we append a single scalar scale parameter.
Thus, \cando searches over a compact transformation space rather than the high-dimensional point coordinates.

\subsection{Search Objective}
\label{subsec:objective}

Optimizing the transform requires a score for each candidate, yet the task loss $\ell$ demands labels we do not have.
Drawing inspiration from the TTA literature, we leverage the model's own output confidence as a self-supervised signal~\citep{tent,memo,liang2020shot}. 
Let $[g(\cdot)]_{i,c}$ represent the predicted probability of class $c$ at point $i$. We maximize the mean maximum probability
\begin{equation}
  J({\btheta}) \;=\; \frac{1}{N}\sum_{i=1}^{N}\,\max_{c}\,\bigl[g(\Tb_{\btheta}(\Pb))\bigr]_{i,c},
  \label{eq:conf}
\end{equation}
which reduces to a single maximum probability for classification. 
This surrogate guides recovery only insofar as it tracks accuracy.
Empirically, the correlation between $J$ and mean Intersection-over-Union (mIoU) across poses landscape, typically ranging between $0.5$ and $0.65$ (see \Cref{sec:dynamics}).

However, indiscriminately maximizing $J$ is hazardous, as an unconstrained search may inadvertently rotate an already-canonical input out of alignment. 
To mitigate this, we introduce a simple yet effective do-no-harm prior that penalizes the transform magnitude and always retains the identity ${\btheta}=\bm{0}$ as a candidate, giving the final selected transform
\begin{equation}
  \hat{\btheta} \;=\; \argmax_{{\btheta}\in\{\bm{0}\}\cup\Theta}\;\bigl[\,J({\btheta})-\lambda\|{\btheta}\|\,\bigr],
  \label{eq:prior}
\end{equation}
where $\Theta$ denotes the pool of candidate transformations evaluated by the optimizer, and $\lambda$ controls the penalty strength. This regularization ensures the input remains untouched unless an alternative pose yields a conclusively higher confidence score, acting as a safeguard for already-upright inputs.

\subsection{The CMA Evolution Strategy}
\label{subsec:cma}

To navigate the set of candidates $\Theta$, the optimizer must operate exclusively on forward evaluations of $J$. 
Given the highly non-convex nature of the pose landscape, we utilize the Covariance Matrix Adaptation Evolution Strategy (CMA-ES)~\citep{hansen2001cmaes,hansen2003reducing}, a population-based evolutionary search well suited for continuous, multi-modal objectives~\citep{sun2022bbt,sun2022bbtv2,park2025zip}.
At each generation $t$, CMA-ES samples a population ${\btheta}_m\sim\mathcal{N}(\bmu_t,\sigma_t^2 \Cb_t)$ defined by its mean $\bmu_t$, step size $\sigma_t$, and covariance $\Cb_t$, then updates these from the ranked scores,
\begin{equation}
\begin{aligned}
  \bmu_{t+1} &= \sum_{m=1}^{\mu} w_m\,{\btheta}_{(m)},\\
  (\sigma_{t+1}, \Cb_{t+1}) &= \mathrm{update}\bigl(\sigma_t, \Cb_t, \{{\btheta}_{(m)}\}\bigr),
\end{aligned}
  \label{eq:cma}
\end{equation}
where ${\btheta}_{(m)}$ are the candidates ordered by $J$ and $w_m$ are rank weights; 
we refer the reader to \citep{hansen2016cma} for the details.

CMA-ES provides two critical advantages for this framework. 
First, its global search capability allows it to bypass local optima that routinely trap gradient-based methods. 
Second, it demonstrates outstanding convergence efficiency among global DFO methods, requiring the fewest queries to locate the canonical pose (see~\Cref{subsec:optimizer} for the details).

Algorithm~\ref{alg:canon} summarizes the complete \cando pipeline. 
Each scored pose ${\btheta}$ returns the model's class-probability output $\Pi_{\btheta}\!=\!g(\Tb_{\btheta}(\Pb))$, from which \Cref{eq:conf} computes the confidence $J({\btheta})$. 
The prediction at the selected pose $\hat{\btheta}$ is cached during the search and inverse-warped to the original frame, so the final output costs no extra query.

\begin{algorithm}[t]
\caption{\cando: Black-Box Pose Canonicalization}
\label{alg:canon}
\begin{algorithmic}[1]
\Require Input point cloud $\Pb$, frozen point cloud model $g$, transform family $\{\Tb_{\btheta}\}$, objective $J$, query budget $M$, population $K$, prior weight $\lambda$, CMA-ES optimizer $\cA$.
\Ensure Prediction for $\Pb$ in its original frame.
\State \textbf{Initialize} evaluation log $\cE \gets \{(\bm{0},\,J(\bm{0}),\,g(\Pb))\}$ with the identity, and query counter $q \gets 1$.
\While{$q < M$}
    \State \textbf{Sample} the next generation of candidates:
    \Statex \qquad $\rhd\;\Theta_q \gets \cA.\operatorname{ask}(\min\{K,\,M-q\})$
    \State \textbf{Evaluate} each candidate with a single query:
    \Statex \qquad $\rhd\;\Pi_{\btheta} \gets g(\Tb_{\btheta}(\Pb)) \quad \forall\,{\btheta}\in\Theta_q$
    \State \textbf{Record} the scores and predictions:
    \Statex \qquad $\rhd\;\cE \gets \cE \cup \{({\btheta},\,J({\btheta}),\,\Pi_{\btheta})\}_{{\btheta}\in\Theta_q}$
    \State \textbf{Update} the optimizer from the ranked scores:
    \Statex \qquad $\rhd\;\cA.\operatorname{tell}(\Theta_q,\,\{J({\btheta})\})$,\quad $q \gets q + |\Theta_q|$
\EndWhile
\State \textbf{Select} the pose via the do-no-harm prior (\Cref{eq:prior}):
\Statex $\rhd\;\hat{\btheta} \gets \argmax_{({\btheta},\,j,\,\Pi)\in\cE} \bigl[\,j-\lambda\|{\btheta}\|\,\bigr]$
\State \Return cached $\hat{\Pi}$ at $\hat{\btheta}$, inverse-warped to original frame.
\end{algorithmic}
\end{algorithm}
\section{Experiments}
\label{sec:evaluations}

\subsection{Experimental Setup}
\label{subsec:setup}

\paragraph{Datasets.}
We evaluate \cando on three standard point cloud benchmarks covering dense and global prediction tasks.
For object part segmentation, we use ShapeNetPart~\citep{yi2016shapenetpart}, which contains CAD objects from 16 categories with point-level part annotations.
For scene semantic segmentation, we use S3DIS~\citep{armeni2016s3dis}, an indoor-scene dataset with dense semantic labels.
For object classification, we use ModelNet40~\citep{wu2015modelnet}, a CAD benchmark with 40 object categories.

\paragraph{Backbone models.}
Our primary evaluation employs PointNet++~\citep{qi2017pointnetpp} and PointNeXt-S~\citep{qian2022pointnext}, two widely adopted backbones representing a foundational and a modern point-based design, both accessed solely through forward queries. 

\paragraph{Implementation details.}
All models are frozen and accessed only through forward queries on a single RTX A6000, and results are averaged over three different random seeds. 
CMA-ES uses two restarts and a population of $10$ to $12$, and headline results take the highest-confidence solution without a magnitude penalty.
Further details are in Supp.~\ref{app:impl}, and the code is available at \url{https://github.com/shpark97/CanDO}.

\paragraph{Baselines.}
We compare \cando with two kinds of methods: gradient-free methods compatible with our black-box setting, and gradient-based methods that require white-box access and serve only as references.
\textit{For gradient-free methods}, we consider three baselines:
\textbf{(1) Principal-Axis Canonicalization}, which rotates the input into the frame of its principal axes, \eg, PCA~\citep{xiao2020pca}, with a confidence-disambiguated variant that resolves the per-axis sign ambiguity by the model's own confidence;
\textbf{(2) Normalization-Statistic Adaptation}, which updates the model's batch-norm statistics per input, \eg, DUA~\citep{mirza2022dua}; 
and \textbf{(3) External-Prior Canonicalization}, scoring candidate poses with a separate frozen classifier instead of the task model's own confidence. 
\textit{For gradient-based methods}, we consider two baselines:
\textbf{(1) Parameter-Space TTA}, minimizing prediction entropy by backpropagating into the model's parameters, \eg, TENT~\citep{tent} and MEMO~\citep{memo};
and \textbf{(2) Gradient-based Canonicalization}, white-box gradient ascent on the input rotation under a matched query budget. 
We additionally report the oracle, a supervised ceiling selecting the best pose by the true metric.

\begin{table}[t]
\centering\small
\caption{Test-time pose recovery performance of frozen point cloud models under pose corruptions across various tasks, scales, and architectures. Relying exclusively on forward queries, \cando bridges a substantial portion of the oracle gap across all evaluated benchmarks and backbones.}
\label{tab:canon}
\vspace{-0.75em}
\setlength{\tabcolsep}{5pt}
\renewcommand{\arraystretch}{1.15}
\resizebox{\linewidth}{!}{%
\begin{tabular}{l c l ccc c}
\toprule
& & & \multicolumn{3}{c}{\textbf{mIoU / Accuracy}} & \\
\cmidrule(lr){4-6}
\textbf{Benchmark} & \textbf{Transform} & \textbf{Backbone} & Corrupted & Recovered & Oracle & \textbf{\% Gap} \\
\midrule
\multirow{4}{*}{ShapeNetPart} & \multirow{2}{*}{Azimuth} & PointNet++  & $0.387{\pm}0.005$ & $\mathbf{0.752}{\pm}0.002$ & $0.869{\pm}0.000$ & \cellcolor{green!27}\textbf{76} \\
                          &                          & PointNeXt-S & $0.396{\pm}0.005$ & $\mathbf{0.718}{\pm}0.003$ & $0.871{\pm}0.000$ & \cellcolor{green!23}\textbf{68} \\
\cmidrule(l){2-7}
                          & \multirow{2}{*}{$SO(3)$} & PointNet++  & $0.352{\pm}0.001$ & $\mathbf{0.602}{\pm}0.003$ & $0.845{\pm}0.001$ & \cellcolor{green!17}\textbf{51} \\
                          &                          & PointNeXt-S & $0.372{\pm}0.002$ & $\mathbf{0.637}{\pm}0.003$ & $0.843{\pm}0.001$ & \cellcolor{green!19}\textbf{56} \\
\midrule
\multirow{2}{*}{S3DIS} & \multirow{2}{*}{Tilt} & PointNet++  & $0.305{\pm}0.005$ & $\mathbf{0.528}{\pm}0.046$ & $0.735{\pm}0.023$ & \cellcolor{green!17}\textbf{52} \\
                       &                       & PointNeXt-S & $0.249{\pm}0.013$ & $\mathbf{0.599}{\pm}0.050$ & $0.756{\pm}0.025$ & \cellcolor{green!23}\textbf{69} \\
\midrule
\multirow{2}{*}{ModelNet40} & \multirow{2}{*}{$SO(3)$} & PointNet++  & $0.125{\pm}0.005$ & $\mathbf{0.340}{\pm}0.007$ & $0.861{\pm}0.003$ & \cellcolor{green!9}\textbf{29} \\
                            &                          & PointNeXt-S & $0.160{\pm}0.004$ & $\mathbf{0.480}{\pm}0.004$ & $0.920{\pm}0.001$ & \cellcolor{green!14}\textbf{42} \\
\bottomrule
\end{tabular}%
}
\end{table}

\paragraph{Evaluation protocol.}
Each test input is corrupted by a random pose, an azimuth, a horizontal-axis tilt, or a random-axis $SO(3)$ rotation~\citep{esteves2018,deng2021vn}. 
As metrics, we report mIoU for segmentation and accuracy for classification, propagating segmentation predictions back to the full cloud by three-nearest-neighbor voting; 
on ShapeNetPart we follow the standard protocol of feeding the object category as a one-hot input and scoring category-masked mIoU.

\begin{table*}[t]
\centering\small
\caption{Performance comparison of pose recovery against established baselines and a supervised oracle, under $SO(3)$ rotation. Using solely forward queries, \cando achieves the highest recovery, outperforming even the white-box baselines. \textbf{Bold} denotes the best results.}
\vspace{-0.5em}
\label{tab:compare}
\resizebox{0.975\linewidth}{!}{%
\begin{tabular}{lcccccc}
\toprule
& & & & \multicolumn{2}{c}{\textbf{mIoU}} & \textbf{Accuracy} \\
\cmidrule(lr){5-6}\cmidrule(lr){7-7}
\textbf{Method} & \textbf{Gradient} & \textbf{Update} & \textbf{Label} & ShapeNetPart~\citep{yi2016shapenetpart} & S3DIS~\citep{armeni2016s3dis} & ModelNet40~\citep{wu2015modelnet} \\
\midrule
Source (No Adaptation)          & \xmark & \xmark & \xmark & $0.351\pm0.004$ & $0.338\pm0.011$ & $0.128\pm0.007$ \\
PCA Canonicalization~\citep{xiao2020pca} & \xmark & \xmark & \xmark & $0.237\pm0.003$ & $0.047\pm0.004$ & $0.221\pm0.008$ \\
PCA + Confidence                & \xmark & \xmark & \xmark & $0.293\pm0.004$ & $0.047\pm0.004$ & $0.260\pm0.009$ \\
External Prior (Classifier)     & \xmark & \xmark & \xmark & $0.359\pm0.005$ & N/A & N/A            \\
DUA~\citep{mirza2022dua}        & \xmark & \cmark & \xmark & $0.356\pm0.005$ & $0.357\pm0.011$ & $0.118\pm0.007$ \\
TENT~\citep{tent}               & \cmark & \cmark & \xmark & $0.353\pm0.005$ & $0.353\pm0.011$ & $0.123\pm0.007$ \\
MEMO~\citep{memo}               & \cmark & \cmark & \xmark & $0.354\pm0.005$ & $0.357\pm0.011$ & $0.126\pm0.007$ \\
Gradient-Based Canonicalization & \cmark & \xmark & \xmark & $0.500\pm0.006$ & $0.440\pm0.017$ & $0.149\pm0.007$ \\
\textbf{Ours (\cando)}                    & \xmark & \xmark & \xmark & $\mathbf{0.606\pm0.006}$ & $\mathbf{0.523\pm0.022}$ & $\mathbf{0.354\pm0.010}$ \\
\midrule
Oracle (Ground Truth)           & \xmark & \xmark & \cmark & $0.845\pm0.003$ & $0.752\pm0.011$ & $0.861\pm0.007$ \\
\bottomrule
\end{tabular}%
}
\end{table*}

\begin{figure}[t]\centering
\vspace{-1em}
\includegraphics[width=0.875\linewidth]{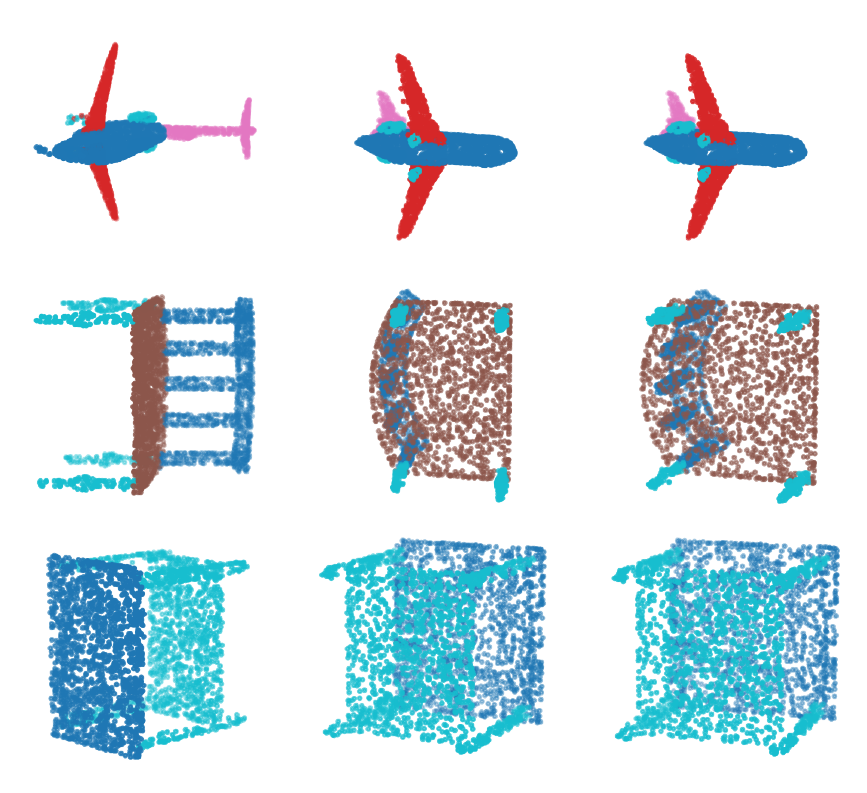}
\caption{Qualitative results of pose recovery on ShapeNetPart. From left to right, each column depicts the misoriented input, the pose recovery achieved by \cando, and the clean upright target.}
\label{fig:canon}
\end{figure}

\subsection{Results}
\label{subsec:canon}

\paragraph{Restoring frozen models under rotation-induced shifts.}
We evaluate whether frozen point cloud models can be recovered from rotation-induced shifts using only forward queries.
As shown in \Cref{tab:canon}, \cando recovers a substantial portion of the oracle gap, \ie, the performance difference between the corrupted input and the oracle-selected transformation.
On PointNet++, it recovers $76\%$ of the gap under azimuth rotations and $51\%$ under arbitrary $SO(3)$ rotations for part segmentation. The corresponding rates are $52\%$ for scene tilt and $29\%$ for $SO(3)$ classification.
For the PointNeXt-S backbone, recovery increases to $68\%$, $56\%$, $69\%$, and $42\%$ across the same settings.
Recovery is comparatively lower for classification because a single softmax confidence correlates less strongly with task accuracy, yet \cando still restores a substantial fraction of the lost performance.

\Cref{fig:canon} illustrates qualitative recovery results on ShapeNet part segmentation. 
The columns denote the misoriented input (left), the pose recovered by \cando (middle), and the clean upright target (right), all color-coded according to their ground-truth parts. 
The recovered orientation closely aligns with the upright target, effectively restoring the accurate per-part labels. 
Collectively, both quantitative and qualitative findings demonstrate that maximizing the inherent confidence of a frozen model successfully guides the input back toward the optimally scoring pose.

\paragraph{Comparison to baselines.}
\label{subsec:compare}

\Cref{tab:compare} compares \cando with representative gradient-free and gradient-based baselines~\citep{xiao2020pca,mirza2022dua,tent,memo} under full $SO(3)$ corruption.
On ShapeNet part segmentation, gradient-free baselines provide little benefit because their scoring signals are weakly aligned with the canonical orientation, reaching at most $0.359$ mIoU.
Gradient-based adaptation performs similarly: entropy-minimization TTA~\citep{tent,memo} remains near $0.35$ mIoU, while even white-box gradient ascent reaches only $0.500$ mIoU after becoming trapped in the non-convex confidence landscape (see \Cref{subsec:optimizer}).
By contrast, \cando achieves $0.606$ mIoU by globally searching model confidence over low-dimensional rotations, outperforming the best gradient-free and white-box gradient baselines by $0.247$ and $0.106$.

The same pattern extends beyond object part segmentation.
On S3DIS under out-of-plane tilt, \cando reaches $0.523$ mIoU. 
Competing baselines remain near the corrupted performance; even white-box gradient ascent reaches only $0.440$ mIoU, while PCA further collapses to $0.047$ by ignoring gravity.
On ModelNet40 classification, \cando reaches $0.354$ accuracy, outperforming the strongest baseline by $0.094$.
Overall, \cando consistently outperforms competing methods across object, scene, and classification tasks while relying solely on forward queries.
\section{Further Analysis}
\label{sec:dynamics}
In this section, we analyze the factors governing \cando's recovery.
We first study transformation recoverability (\Cref{subsec:auggap}), showing that recovery requires both an augmentation gap and an accuracy-aligned confidence surrogate.
We then characterize the confidence landscape and compare optimization strategies (\Cref{subsec:optimizer}), highlighting the need for global search.
Finally, we evaluate surrogate reliability (\Cref{subsec:gameable}), the identity prior (\Cref{subsec:reach}), and equivariant backbones (\Cref{subsec:equiv}) to determine when confidence-only recovery is sufficient, insufficient, or unnecessary.

\subsection{Transformation Recoverability}
\label{subsec:auggap}
To determine which transformations are recoverable, we first probe a frozen PointNet++ model trained with yaw augmentation on S3DIS.
We evaluate five transformations that differ in whether they were covered during training and whether confidence remains aligned with accuracy.
As shown in \Cref{tab:auggap}, successful recovery requires two conditions: (i) an augmentation gap, \ie, a transformation that is not sufficiently covered during training and degrades performance, and (ii) a confidence surrogate that remains informative under the transformation.
Yaw has almost no recoverable headroom because the model is already robust to it through training augmentation.
Scale and color also produce small accuracy drops of $0.08$ and $0.06$, leaving limited room for improvement.
A vertical offset causes a larger accuracy drop of $0.22$, but remains difficult to recover because the confidence objective does not select offsets that improve task accuracy.
In contrast, unseen out-of-plane tilt satisfies both conditions and is recovered by \cando, improving accuracy by $0.22$.
We further validate this behavior on alternative point cloud architectures in Supp.~\ref{app:auggap}.

The preceding analysis is based on existing training configurations and is therefore correlational.
To test the causal role of augmentation gaps, we train PointNet++ models on ShapeNetPart under four rotation augmentation regimes: none, yaw, tilt, and full $SO(3)$.
We then apply the corresponding test-time rotations and measure corrupted-to-recovered mIoU.
As shown in \Cref{tab:augablation}, recovery is largest when the test corruption is poorly covered by the training augmentation.
For example, the model trained without rotation augmentation improves substantially under yaw, tilt, and $SO(3)$ corruptions.
The yaw-augmented model shows little change under yaw but larger gains under unseen tilt and $SO(3)$ rotations.
Conversely, models trained with tilt or full $SO(3)$ augmentation show smaller recovery because their corrupted performance is already closer to clean performance.
These results support the augmentation-gap hypothesis while also showing that coverage is not binary: partial gains can remain when training augmentation only partially matches the test rotation.

\begin{table}[t]
\centering\small
\caption{Conditions for recoverability on the S3DIS PointNet++ model. Recovery requires performance headroom (an accuracy drop) and an accuracy-aligned confidence surrogate. The unseen tilt (green) best satisfies both prerequisites. Recovery $\Delta$ denotes the accuracy change after label-free recovery.}
\vspace{-0.75em}
\label{tab:auggap}
\resizebox{\linewidth}{!}{%
\begin{tabular}{ccccc}
\toprule
\textbf{Transform} & \textbf{Augmented} & \textbf{Accuracy Drop} & \textbf{Recovery} $\Delta$ & \textbf{Recoverable} \\
\midrule
Yaw        & Yes & 0.01 & $+0.00$ & \xmark~(no headroom) \\
Scale      & No  & 0.08 & $+0.00$ & \xmark~(no headroom) \\
\rowcolor{red!8}
$z$-offset & No  & 0.22 & $-0.02$ & \xmark~(misaligned confidence) \\
Color      & No  & 0.06 & $+0.03$ & \xmark~(no headroom) \\
\rowcolor{green!12}
Tilt       & No  & 0.42 & $+0.22$ & \cmark \\
\bottomrule
\end{tabular}%
}
\end{table}

\begin{table}[t]\centering\small
\caption{Causal link between augmentation gaps and recoverability on ShapeNetPart.
Cells report corrupted $\to$ recovered mIoU, shaded by recovery gain so that darker red indicates larger improvements.
Recovery is largest when the corruption is absent from training augmentation and reduced when it is covered.}
\vspace{-0.75em}
\label{tab:augablation}
\resizebox{\linewidth}{!}{%
\begin{tabular}{ccccc}
\toprule
\textbf{Augmentation} & \textbf{Clean} & Yaw & Tilt & $SO(3)$ \\
\midrule
None    & 0.828 & \cellcolor{red!44}$0.390{\to}0.562$ & \cellcolor{red!34}$0.505{\to}0.613$ & \cellcolor{red!34}$0.447{\to}0.558$ \\
Yaw     & 0.825 & \cellcolor{red!6}$0.824{\to}0.829$ & \cellcolor{red!34}$0.529{\to}0.655$ & \cellcolor{red!34}$0.539{\to}0.668$ \\
Tilt    & 0.826 & \cellcolor{red!14}$0.733{\to}0.765$ & \cellcolor{red!6}$0.813{\to}0.819$ & \cellcolor{red!14}$0.779{\to}0.798$ \\
$SO(3)$ & 0.820 & \cellcolor{red!6}$0.804{\to}0.811$ & \cellcolor{red!6}$0.808{\to}0.815$ & \cellcolor{red!6}$0.805{\to}0.812$ \\
\bottomrule
\end{tabular}%
}
\end{table}

\subsection{Optimization Strategy}
\label{subsec:optimizer}

\begin{figure*}[t]\centering
\includegraphics[width=0.9\textwidth]{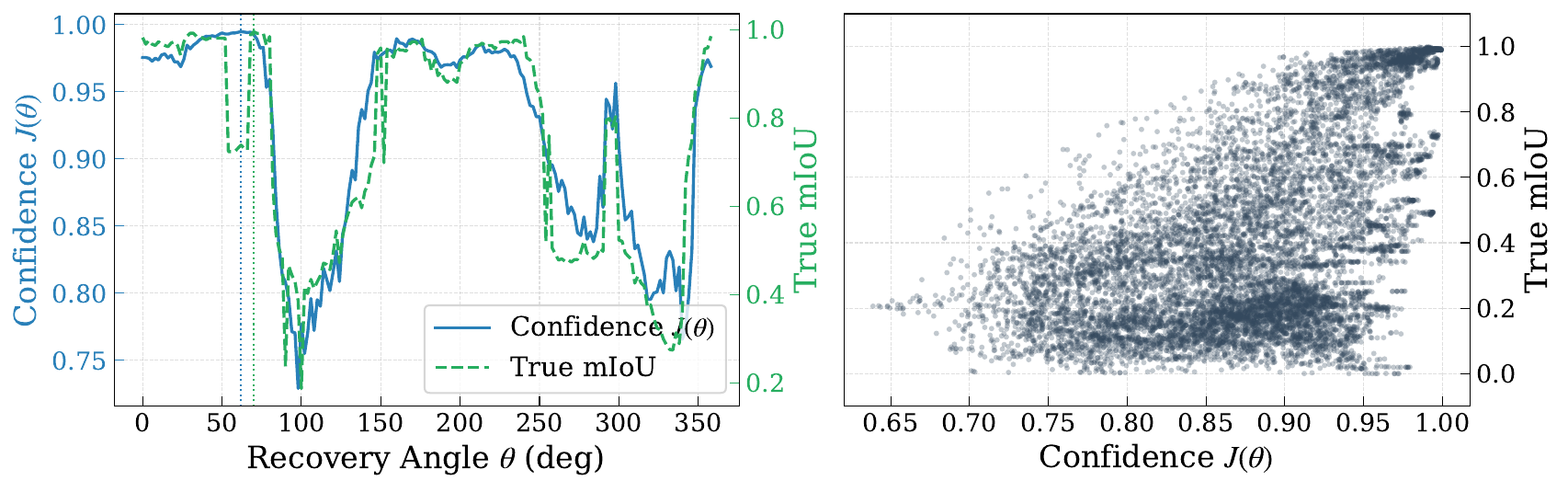}
\vspace{-0.5em}
\caption{(Left) For one shape, the confidence objective is multi-modal yet peaks at the same pose as the
true mIoU, so a local search is trapped while a global one is not. (Right) Across shapes and poses the
objective tracks accuracy, with a pooled Pearson correlation of $0.51$.}
\label{fig:landscape}
\end{figure*}

We next study recovery from an optimization perspective.
To characterize the objective optimized by forward-query methods, we sweep the recovery rotation over corrupted inputs and record the confidence objective $J(\theta)$ alongside accuracy.
As shown in \Cref{fig:landscape}, the resulting landscape is highly multi-modal but remains well aligned with task performance.
Although $J(\theta)$ contains multiple local maxima for each shape, its global maximum consistently coincides with the pose achieving the highest mIoU.
Local optimizers therefore become trapped in suboptimal peaks, whereas global search reliably identifies the correct pose.
Across shapes and rotations, $J(\theta)$ exhibits a pooled Pearson correlation of $0.51$ with accuracy.
These results indicate that the confidence objective is informative, making landscape multi-modality the main challenge for successful recovery.

We then compare representative DFO methods on the full $SO(3)$ part-segmentation recovery task using a matched budget of approximately $300$ queries per shape.
As detailed in \Cref{tab:est}, performance sharply diverges between local and global methods. 
Four local gradient-estimation methods~\citep{conn2009introduction, wierstra2014nes, williams1992reinforce, spall1992spsa} plateau, recovering only $17\%$ to $27\%$ of the oracle performance gap. 
By contrast, global search algorithms~\citep{shahriari2015taking, bergstra2012random, hansen2001cmaes} bypass these spurious local optima and recover roughly twice as much: random search reaches $53\%$, CMA-ES $55\%$, and Bayesian optimization $57\%$. 
This local-global gap reinforces that navigating the multi-modal landscape is the critical determinant of successful recovery.

Having established the advantage of global search, we next compare the query efficiency of the three global optimizers.
As shown in \Cref{tab:bo}, performance depends on the query budget.
CMA-ES dominates under constrained budgets, recovering $52\%$ and $54\%$ of the gap with $60$ and $120$ queries, whereas random search and Bayesian optimization recover $42\%$ and $41\%$ at $60$ queries.
The methods converge after several hundred queries, where Bayesian optimization reaches $56\%$ at $240$.
Because practical deployment favors low query budgets, we adopt CMA-ES as the default optimizer.

\begin{table}[t]\centering\small
\caption{Optimizer comparison on $SO(3)$ part segmentation recovery. Performance is measured as the percentage of the oracle gap recovered under a matched budget of approximately $300$ queries per shape. Global search strategies significantly outperform local estimators, which plateau at spurious local optima.}
\vspace{-0.75em}
\label{tab:est}
\resizebox{\linewidth}{!}{%
\begin{tabular}{llc}
\toprule
\textbf{Optimizer} & \textbf{Family} & \textbf{\% of Oracle Gap} \\
\midrule
Coordinate Finite Differences~\citep{conn2009introduction} & Local  & 17 \\
NES~\citep{wierstra2014nes}   & Local  & 18 \\
Score-Function~\citep{williams1992reinforce} & Local  & 19 \\
SPSA~\citep{spall1992spsa}    & Local  & 27 \\
\rowcolor{green!8}
Bayesian Optimization~\citep{shahriari2015taking} & Global & \textbf{57} \\
\rowcolor{green!8}
Random Search~\citep{bergstra2012random} & Global & 53 \\
\rowcolor{green!8}
CMA-ES~\citep{hansen2001cmaes} & Global & 55 \\
\bottomrule
\end{tabular}%
}
\end{table}

\begin{table}[t]\centering\small
\caption{Query efficiency of global search algorithms on $SO(3)$ recovery. Performance is reported as the percentage of the oracle gap recovered across varying query budgets. Algorithmic rankings are budget-dependent: CMA-ES dominates constrained regimes, whereas Bayesian optimization prevails at higher query counts.}
\vspace{-0.75em}
\label{tab:bo}
\resizebox{0.9\linewidth}{!}{%
\begin{tabular}{cccc}
\toprule
\textbf{Budget} & \textbf{Bayesian}~\citep{shahriari2015taking} & \textbf{Random Search}~\citep{bergstra2012random} & \textbf{CMA-ES}~\citep{hansen2001cmaes} \\
\midrule
60  & 41 & 42 & \textbf{52} \\
120 & 46 & 43 & \textbf{54} \\
240 & \textbf{56} & 51 & 53 \\
\bottomrule
\end{tabular}%
}
\end{table}

\subsection{Surrogate Reliability}
\label{subsec:gameable}
We next assess whether the confidence surrogate remains reliable beyond pure rotation shifts.
This matters because \cando selects transformations by maximizing confidence, which can fail if confidence increases without improving task accuracy.
We therefore evaluate the recovery procedure on the PointCloud-C corruption suite~\citep{pointcloudc}, including transform, noise, and density corruptions.
As shown in \Cref{tab:corr}, dense per-point segmentation scores are more reliable than a single classification softmax.
The segmentation objective remains non-negative across all corruption families and achieves a mean gain of $+0.002$.
It also keeps the identity transformation on $89\%$ of inputs, avoiding unnecessary changes when no beneficial pose exists.
By contrast, the classification softmax gains only $+0.015$ on transform corruptions but loses $0.070$ on noise and $0.028$ on density.
Its mean change is $-0.034$, with the identity retained on only $62\%$ of inputs.
These results suggest that surrogate reliability improves with denser confidence signals.

\begin{table}[t]\centering\small
\caption{Robustness across PointCloud-C corruptions. The dense per-point segmentation objective remains stable under confidence maximization, whereas the single classification softmax is more susceptible to overconfidence and performance degradation.}
\vspace{-0.75em}
\label{tab:corr}
\resizebox{\linewidth}{!}{%
\begin{tabular}{lccccc}
\toprule
\textbf{Objective} & \textbf{Transform} $\Delta$ & \textbf{Noise} $\Delta$ & \textbf{Density} $\Delta$ & \textbf{Mean} $\Delta$ & \textbf{Identity Kept} \\
\midrule
Part Segmentation & \cellcolor{green!10}$+0.001$ & \cellcolor{green!10}$+0.002$ & \cellcolor{green!10}$+0.001$ & \cellcolor{green!10}$+0.002$ & $89\%$ \\
Classification    & \cellcolor{green!12}$+0.015$ & \cellcolor{red!14}$-0.070$ & \cellcolor{red!10}$-0.028$ & \cellcolor{red!10}$-0.034$ & $62\%$ \\
\bottomrule
\end{tabular}%
}
\end{table}

To understand where this difference comes from, \Cref{tab:appmc} provides a per-corruption breakdown.
The gains mainly come from rotation, which improves by $0.055$.
Non-pose corruptions instead show mild drops because confidence can increase without improving correctness.
This effect is largest for global outlier additions, where accuracy falls by $0.093$.
Replacing confidence with a marginal-entropy objective~\citep{memo} amplifies the same pattern.
It improves rotation recovery to $+0.169$ but worsens the mean change from $-0.034$ to $-0.058$.
These results suggest that the reliability of confidence-only recovery depends on the information contained in the surrogate.
For corruptions that induce overconfidence without a correctable pose, reliable recovery may require additional task-specific constraints or external priors from diffusion or foundation models~\citep{cloudfixer,focal}.

\begin{table}[t]\centering\small
\caption{Change in accuracy ($\Delta$) across ModelNet-C corruptions when optimized using confidence and marginal-entropy objectives. Performance gains are predominantly localized to rotational transformations, and the marginal-entropy formulation fails to mitigate the performance degradation observed under non-pose corruptions.}
\vspace{-0.75em}
\label{tab:appmc}
\resizebox{\linewidth}{!}{%
\begin{tabular}{llcc}
\toprule
\textbf{Corruption} & \textbf{Family} & $\Delta$ \textbf{Confidence} & $\Delta$ \textbf{Marginal Entropy} \\
\midrule
Rotate          & Transform & \cellcolor{green!12}$+0.055$ & \cellcolor{green!12}$+0.169$ \\
Scale           & Transform & \cellcolor{red!8}$-0.026$ & \cellcolor{red!8}$-0.037$ \\
Jitter          & Noise     & \cellcolor{red!8}$-0.054$ & \cellcolor{red!12}$-0.125$ \\
Add (Global)    & Noise     & \cellcolor{red!12}$-0.093$ & \cellcolor{red!12}$-0.150$ \\
Add (Local)     & Noise     & \cellcolor{red!8}$-0.064$ & \cellcolor{red!12}$-0.150$ \\
Dropout (Global)& Density   & \cellcolor{red!4}$-0.018$ & \cellcolor{red!8}$-0.038$ \\
Dropout (Local) & Density   & \cellcolor{red!8}$-0.038$ & \cellcolor{red!8}$-0.050$ \\
\midrule
Mean Over Seven & ---       & $-0.034$ & $-0.058$ \\
\bottomrule
\end{tabular}%
}
\end{table}

\begin{figure}[t]\centering
\begin{subfigure}{0.49\linewidth}\centering
\includegraphics[width=\linewidth]{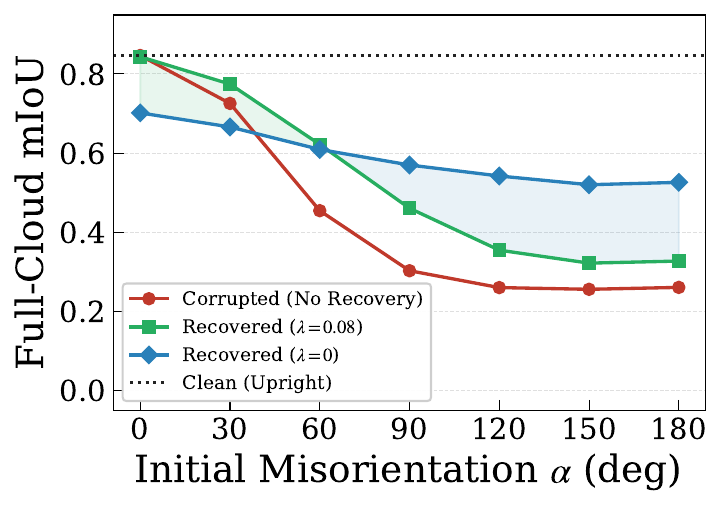}
\caption{ShapeNet}
\label{fig:recangle_a}
\end{subfigure}
\begin{subfigure}{0.49\linewidth}\centering
\includegraphics[width=\linewidth]{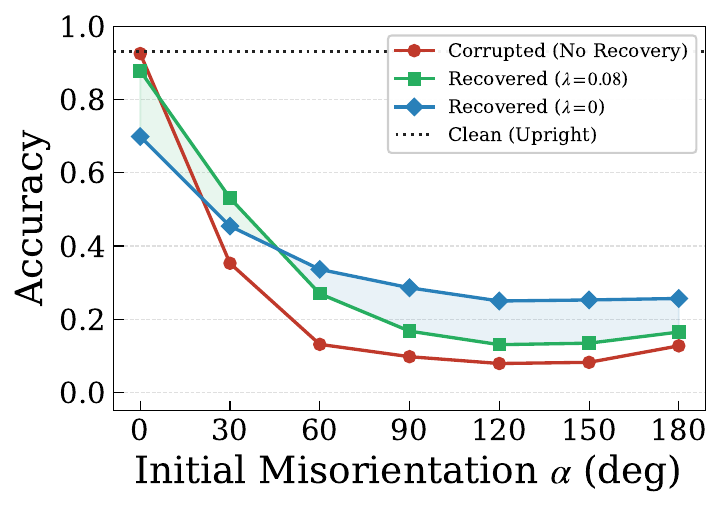}
\caption{ModelNet40}
\label{fig:recangle_c}
\end{subfigure}
\vspace{-0.5em}
\caption{Recovery across initial misorientation angles $\alpha$ under random-axis $SO(3)$ rotations on ShapeNetPart and ModelNet40. The identity prior preserves canonical inputs but struggles with near-inversions. Conversely, omitting the prior improves large-rotation recovery but degrades well-posed inputs.}
\label{fig:recangle}
\end{figure}

\subsection{Identity Prior}
\label{subsec:reach}

We next examine the stability-recovery trade-off introduced by the identity prior.
We evaluate the frozen part-segmentation and classification models across initial misorientation angles $\alpha$ with and without the prior.
As shown in \Cref{fig:recangle}, the default prior $\lambda{=}0.08$ performs best at small-to-moderate misorientations.
It preserves canonical inputs and stays within $0.003$ of clean accuracy at $\alpha{=}0$.
However, the magnitude penalty discourages the large corrective rotations needed for near-inversions.
Removing the prior with $\lambda{=}0$ reverses this behavior.
It recovers roughly half of the oracle gap even at $180^\circ$ but harms well-posed inputs.
At $\alpha{=}0$, accuracy drops from $0.847$ to $0.702$.
The two settings cross between $60^\circ$ and $90^\circ$.
Since our target deployment setting involves moderate rotational offsets, we use $\lambda{=}0.08$ by default.
Deployments that expect extreme misorientations may benefit from a smaller $\lambda$.

\subsection{Beyond Equivariance}
\label{subsec:equiv}

We close the analysis by asking whether architectural rotation robustness removes the recovery headroom exploited by \cando.
To this end, we evaluate frozen equivariant part-segmentation backbones based on Vector Neurons~\citep{deng2021vn}.
Because these networks use rotation-equivariant operations, pose corruptions rotate the feature space without changing per-point predictions.
As shown in \Cref{tab:vnseg}, VN backbones remain invariant under $SO(3)$.
Their corrupted and \cando mIoU therefore match clean performance, making \cando effectively an identity operation.
In contrast, non-equivariant PointNet++ and PointNeXt-S degrade under $SO(3)$ rotations, and \cando recovers a substantial portion of the gap.

This identity behavior is a useful feature rather than a limitation.
First, \cando does not degrade rotation-robust models, enabling a unified deployment policy without prior knowledge of backbone equivariance.
Second, architectural equivariance can lower the clean-performance ceiling, causing VN backbones to underperform non-equivariant counterparts on canonical inputs.
This gap is larger for object-level classification.
Third, recent embodied agents~\citep{huang2024embodied}, robotic manipulation policies~\citep{chen2023polarnet}, and grasping models~\citep{guo2024phygrasp,wu2026hug} often build 3D perception front-ends from frozen non-equivariant encoders such as PointNet++ and PointNeXt.
These front-ends encounter arbitrary object poses but cannot be retrained, making post-deployment recovery necessary when equivariant redesign is unavailable.

\begin{table}[t]\centering\small
\caption{Performance of frozen ShapeNet models under $SO(3)$ corruption. Equivariant Vector-Neurons~\citep{deng2021vn} remain strictly invariant, rendering \cando an identity operation (green). Conversely, non-equivariant backbones suffer severe degradation, from which \cando recovers a substantial portion of the oracle gap (red).}
\label{tab:vnseg}
\vspace{-0.75em}
\resizebox{\linewidth}{!}{%
\begin{tabular}{lcccc}
\toprule
\textbf{Frozen model} & \textbf{Equivariance} & \textbf{Clean} & $SO(3)$ & $+$\cando \\
\midrule
PointNet++~\citep{qi2017pointnetpp}    & \xmark & 0.847 & \cellcolor{red!16}0.352 & \cellcolor{red!16}\textbf{0.602} \\
PointNeXt-S~\citep{qian2022pointnext}  & \xmark & 0.841 & \cellcolor{red!16}0.372 & \cellcolor{red!16}\textbf{0.637} \\
VN-PointNet~\citep{deng2021vn}         & \cmark & 0.794 & \cellcolor{green!12}0.794 & \cellcolor{green!12}0.794 \\
VN-DGCNN~\citep{deng2021vn}            & \cmark & 0.774 & \cellcolor{green!12}0.774 & \cellcolor{green!12}0.774 \\
\bottomrule
\end{tabular}%
}
\end{table}
\section{Conclusion}
\label{sec:conclusion}

This work has studied score-only adaptation for frozen 3D point cloud models under rotation-induced distribution shifts.
We showed that recovery is possible without weights, gradients, or labels by casting test-time canonicalization as a compact search over input rotations.
Our method, \cando, uses DFO to maximize a confidence surrogate over this low-dimensional space.
An identity prior prevents unnecessary changes when the input is already well aligned.
Across tasks, scales, and architectures, \cando recovers a substantial fraction of the performance lost under unseen rotations using only forward queries.
Our analysis identifies the key factors enabling recovery: an augmentation gap, confidence-accuracy alignment, and global search over a multi-modal landscape.
It also clarifies when confidence-only adaptation is insufficient.
These findings provide a practical path toward robust black-box 3D perception and motivate future methods that combine forward-query pose recovery with task-specific constraints or external generative priors.

\section*{Acknowledgements}
This work was supported by the Institute of Information \& Communications Technology Planning \& Evaluation (IITP) grant funded by the Korea government (MSIT) (No.~RS-2025-25441838 and No.~RS-2022-II220926).

{
    \small
    \bibliographystyle{ieeenat_fullname}
    \bibliography{main}

@String(CVPR= {IEEE Conf. Comput. Vis. Pattern Recog.})

@String(ICCV= {Int. Conf. Comput. Vis.})

@String(ECCV= {Eur. Conf. Comput. Vis.})

@String(ICLR = {Int. Conf. Learn. Represent.})

@String(CVPR  = {CVPR})

@String(ICCV  = {ICCV})

@String(ECCV  = {ECCV})

@String(ICLR  = {ICLR})

@inproceedings{cloudfixer,
  title={Cloudfixer: Test-time adaptation for 3d point clouds via diffusion-guided geometric transformation},
  author={Shim, Hajin and Kim, Changhun and Yang, Eunho},
  booktitle=ECCV,
  year={2024}
}

@inproceedings{focal,
  title={Test-time canonicalization by foundation models for robust perception},
  author={Singhal, Utkarsh and Feng, Ryan and Yu, Stella X and Prakash, Atul},
  booktitle={ICML},
  year={2025}
}

@inproceedings{mate,
  title={Mate: Masked autoencoders are online 3d test-time learners},
  author    = {Mirza, M. Jehanzeb and Shin, Inkyu and Lin, Wei and Schriebl, Andreas and Sun, Kunyang and
                 Choe, Jaesung and Kozinski, Mateusz and Possegger, Horst and Kweon, In So and Yoon, Kun-Jin and Bischof, Horst},
  booktitle=ICCV,
  year={2023}
}

@inproceedings{memo,
  title={Memo: Test time robustness via adaptation and augmentation},
  author={Zhang, Marvin and Levine, Sergey and Finn, Chelsea},
  booktitle={NeurIPS},
  year={2022}
}

@inproceedings{tent,
  title={Tent: Fully test-time adaptation by entropy minimization},
  author={Wang, Dequan and Shelhamer, Evan and Liu, Shaoteng and Olshausen, Bruno and Darrell, Trevor},
  booktitle={ICLR},
  year={2021}
}

@inproceedings{kaba,
  title={Equivariance with learned canonicalization functions},
  author={Kaba, S{\'e}kou-Oumar and Mondal, Arnab Kumar and Zhang, Yan and Bengio, Yoshua and Ravanbakhsh, Siamak},
  booktitle={ICML},
  year={2023}
}

@inproceedings{sun2021canonical,
  title={Canonical capsules: Self-supervised capsules in canonical pose},
  author={Sun, Weiwei and Tagliasacchi, Andrea and Deng, Boyang and Sabour, Sara and Yazdani, Soroosh and Hinton, Geoffrey E and Yi, Kwang Moo},
  booktitle={NeurIPS},
  year={2021}
}

@inproceedings{sajnani2022condor,
  title={Condor: Self-supervised canonicalization of 3d pose for partial shapes},
  author={Sajnani, Rahul and Poulenard, Adrien and Jain, Jivitesh and Dua, Radhika and Guibas, Leonidas J and Sridhar, Srinath},
  booktitle=CVPR,
  year={2022}
}

@inproceedings{spezialetti2020compass,
  title={Learning to orient surfaces by self-supervised spherical cnns},
  author={Spezialetti, Riccardo and Stella, Federico and Marcon, Marlon and Silva, Luciano and Salti, Samuele and Di Stefano, Luigi},
  booktitle={NeurIPS},
  year={2020}
}

@inproceedings{pointcloudc,
  title={Benchmarking and analyzing point cloud classification under corruptions},
  author={Ren, Jiawei and Pan, Liang and Liu, Ziwei},
  booktitle={ICML},
  year={2022}
}

@inproceedings{qi2017pointnet,
  title={Pointnet: Deep learning on point sets for 3d classification and segmentation},
  author={Qi, Charles R and Su, Hao and Mo, Kaichun and Guibas, Leonidas J},
  booktitle=CVPR,
  year={2017}
}

@inproceedings{qi2017pointnetpp,
  title={Pointnet++: Deep hierarchical feature learning on point sets in a metric space},
  author={Qi, Charles Ruizhongtai and Yi, Li and Su, Hao and Guibas, Leonidas},
  booktitle={NeurIPS},
  year={2017}
}

@inproceedings{qian2022pointnext,
  title={Pointnext: Revisiting pointnet++ with improved training and scaling strategies},
  author={Qian, Guocheng and Li, Yuchen and Peng, Houwen and Mai, Jinjie and Hammoud, Hasan and Elhoseiny, Mohamed and Ghanem, Bernard},
  booktitle={NeurIPS},
  year={2022}
}

@article{wang2019dgcnn,
  title={Dynamic graph cnn for learning on point clouds},
  author={Wang, Yue and Sun, Yongbin and Liu, Ziwei and Sarma, Sanjay E and Bronstein, Michael M and Solomon, Justin M},
  journal={ACM Transactions on Graphics},
  year={2019}
}

@inproceedings{deng2021vn,
  title={Vector neurons: A general framework for so (3)-equivariant networks},
  author={Deng, Congyue and Litany, Or and Duan, Yueqi and Poulenard, Adrien and Tagliasacchi, Andrea and Guibas, Leonidas J},
  booktitle=ICCV,
  year={2021}
}

@inproceedings{deng2022pointvector,
  title={Pointvector: A vector representation in point cloud analysis},
  author={Deng, Xin and Zhang, WenYu and Ding, Qing and Zhang, XinMing},
  booktitle=CVPR,
  year={2023}
}

@inproceedings{pang2022pointmae,
  title={Masked autoencoders for point cloud self-supervised learning},
  author={Pang, Yatian and Tay, Eng Hock Francis and Yuan, Li and Chen, Zhenghua},
  booktitle=ECCV,
  year={2022}
}

@inproceedings{zhang2019riconv,
  title={Rotation invariant convolutions for 3d point clouds deep learning},
  author={Zhang, Zhiyuan and Hua, Binh-Son and Rosen, David W and Yeung, Sai-Kit},
  booktitle={3DV},
  year={2019}
}

@inproceedings{wu2015modelnet,
  title={3d shapenets: A deep representation for volumetric shapes},
  author={Wu, Zhirong and Song, Shuran and Khosla, Aditya and Yu, Fisher and Zhang, Linguang and Tang, Xiaoou and Xiao, Jianxiong},
  booktitle=CVPR,
  year={2015}
}

@article{yi2016shapenetpart,
  title={A scalable active framework for region annotation in 3d shape collections},
  author={Yi, Li and Kim, Vladimir G and Ceylan, Duygu and Shen, I-Chao and Yan, Mengyan and Su, Hao and Lu, Cewu and Huang, Qixing and Sheffer, Alla and Guibas, Leonidas},
  journal={ACM Transactions on Graphics},
  year={2016}
}

@inproceedings{armeni2016s3dis,
  title={3d semantic parsing of large-scale indoor spaces},
  author={Armeni, Iro and Sener, Ozan and Zamir, Amir R and Jiang, Helen and Brilakis, Ioannis and Fischer, Martin and Savarese, Silvio},
  booktitle=CVPR,
  year={2016}
}

@inproceedings{uy2019scanobjectnn,
  title={Revisiting point cloud classification: A new benchmark dataset and classification model on real-world data},
  author={Uy, Mikaela Angelina and Pham, Quang-Hieu and Hua, Binh-Son and Nguyen, Thanh and Yeung, Sai-Kit},
  booktitle=ICCV,
  year={2019}
}

@article{spall1992spsa,
  title={Multivariate stochastic approximation using a simultaneous perturbation gradient approximation},
  author={Spall, James C},
  journal={IEEE transactions on automatic control},
  year={1992}
}

@article{wierstra2014nes,
  title={Natural evolution strategies},
  author={Wierstra, Daan and Schaul, Tom and Glasmachers, Tobias and Sun, Yi and Peters, Jan and Schmidhuber, J{\"u}rgen},
  journal={JMLR},
  year={2014}
}

@article{hansen2001cmaes,
  title={Completely derandomized self-adaptation in evolution strategies},
  author={Hansen, Nikolaus and Ostermeier, Andreas},
  journal={Evolutionary computation},
  year={2001}
}

@article{salimans2017es,
  title={Evolution strategies as a scalable alternative to reinforcement learning},
  author={Salimans, Tim and Ho, Jonathan and Chen, Xi and Sidor, Szymon and Sutskever, Ilya},
  journal={arXiv preprint arXiv:1703.03864},
  year={2017}
}

@article{bergstra2012random,
  title={Random search for hyper-parameter optimization.},
  author={Bergstra, James and Bengio, Yoshua},
  journal=JMLR,
  year={2012}
}

@article{liu2020zoo,
  title={A primer on zeroth-order optimization in signal processing and machine learning: Principals, recent advances, and applications},
  author={Liu, Sijia and Chen, Pin-Yu and Kailkhura, Bhavya and Zhang, Gaoyuan and Hero III, Alfred O and Varshney, Pramod K},
  journal={IEEE Signal Processing Magazine},
  year={2020}
}

@article{williams1992reinforce,
  title={Simple statistical gradient-following algorithms for connectionist reinforcement learning},
  author={Williams, Ronald J},
  journal={Machine learning},
  year={1992}
}

@article{rodrigues1840,
  title={Des lois g{\'e}om{\'e}triques qui r{\'e}gissent les d{\'e}placements d'un syst{\`e}me solide dans l'espace, et de la variation des coordonn{\'e}es provenant de ces d{\'e}placements consid{\'e}r{\'e}s ind{\'e}pendamment des causes qui peuvent les produire},
  author={Rodrigues, Olinde},
  journal={Journal de math{\'e}matiques pures et appliqu{\'e}es},
  year={1840}
}

@book{murray1994mathematical,
  title={A mathematical introduction to robotic manipulation},
  author={Li, Zexiang and Sastry, Shankar and Murray, R},
  year={2017},
  publisher={CRC press}
}

@inproceedings{shoemake1985,
  title={Animating rotation with quaternion curves},
  author={Shoemake, Ken},
  booktitle={Proceedings of the 12th annual conference on Computer graphics and interactive techniques},
  year={1985}
}

@article{guo2021survey,
  title={Deep learning for 3d point clouds: A survey},
  author={Guo, Yulan and Wang, Hanyun and Hu, Qingyong and Liu, Hao and Liu, Li and Bennamoun, Mohammed},
  journal={TPAMI},
  year={2020}
}

@inproceedings{thomas2019kpconv,
  title={Kpconv: Flexible and deformable convolution for point clouds},
  author={Thomas, Hugues and Qi, Charles R and Deschaud, Jean-Emmanuel and Marcotegui, Beatriz and Goulette, Fran{\c{c}}ois and Guibas, Leonidas J},
  booktitle=ICCV,
  year={2019}
}

@inproceedings{zhao2021pointtransformer,
  title={Point transformer},
  author={Zhao, Hengshuang and Jiang, Li and Jia, Jiaya and Torr, Philip HS and Koltun, Vladlen},
  booktitle=ICCV,
  year={2021}
}

@inproceedings{geiger2012kitti, 
  title={Are we ready for autonomous driving? the kitti vision benchmark suite},
  author={Geiger, Andreas and Lenz, Philip and Urtasun, Raquel},
  booktitle=CVPR, 
  year={2012}
}

@inproceedings{caesar2020nuscenes,
  title={nuscenes: A multimodal dataset for autonomous driving},
  author={Caesar, Holger and Bankiti, Varun and Lang, Alex H and Vora, Sourabh and Liong, Venice Erin and Xu, Qiang and Krishnan, Anush and Pan, Yu and Baldan, Giancarlo and Beijbom, Oscar},
  booktitle=CVPR,
  year={2020},
}

@inproceedings{tramer2016stealing,
  title={Stealing machine learning models via prediction $\{$APIs$\}$},
  author={Tram{\`e}r, Florian and Zhang, Fan and Juels, Ari and Reiter, Michael K and Ristenpart, Thomas},
  booktitle={USENIX Security Symposium},
  year={2016}
}

@inproceedings{chen2017zoo,
  title={Zoo: Zeroth order optimization based black-box attacks to deep neural networks without training substitute models},
  author={Chen, Pin-Yu and Zhang, Huan and Sharma, Yash and Yi, Jinfeng and Hsieh, Cho-Jui},
  booktitle={ACM workshop on artificial intelligence and security},
  year={2017}
}

@inproceedings{ilyas2018blackbox,
  title={Black-box adversarial attacks with limited queries and information},
  author={Ilyas, Andrew and Engstrom, Logan and Athalye, Anish and Lin, Jessy},
  booktitle={ICML},
  year={2018}
}

@inproceedings{sun2020ttt,
  title={Test-time training with self-supervision for generalization under distribution shifts},
  author={Sun, Yu and Wang, Xiaolong and Liu, Zhuang and Miller, John and Efros, Alexei and Hardt, Moritz},
  booktitle={ICML},
  year={2020}
}

@article{thomas2018tfn,
  title={Tensor field networks: Rotation-and translation-equivariant neural networks for 3d point clouds},
  author={Thomas, Nathaniel and Smidt, Tess and Kearnes, Steven and Yang, Lusann and Li, Li and Kohlhoff, Kai and Riley, Patrick},
  journal={arXiv preprint arXiv:1802.08219},
  year={2018}
}

@inproceedings{esteves2018,
  title={Learning so (3) equivariant representations with spherical cnns},
  author={Esteves, Carlos and Allen-Blanchette, Christine and Makadia, Ameesh and Daniilidis, Kostas},
  booktitle=ECCV,
  year={2018}
}

@inproceedings{liang2020shot,
  title={Do we really need to access the source data? source hypothesis transfer for unsupervised domain adaptation},
  author={Liang, Jian and Hu, Dapeng and Feng, Jiashi},
  booktitle={ICML},
  year={2020}
}

@inproceedings{mirza2022dua,
  title={The norm must go on: Dynamic unsupervised domain adaptation by normalization},
  author={Mirza, M Jehanzeb and Micorek, Jakub and Possegger, Horst and Bischof, Horst},
  booktitle=CVPR,
  year={2022}
}

@inproceedings{wang2024bftt3d,
  title={Backpropagation-free network for 3d test-time adaptation},
  author={Wang, Yanshuo and Cheraghian, Ali and Hayder, Zeeshan and Hong, Jie and Ramasinghe, Sameera and Rahman, Shafin and Ahmedt-Aristizabal, David and Li, Xuesong and Petersson, Lars and Harandi, Mehrtash},
  booktitle=CVPR,
  year={2024}
}

@inproceedings{huang2024embodied,
  title={An embodied generalist agent in 3d world},
  author={Huang, Jiangyong and Yong, Silong and Ma, Xiaojian and Linghu, Xiongkun and Li, Puhao and Wang, Yan and Li, Qing and Zhu, Song-Chun and Jia, Baoxiong and Huang, Siyuan},
  booktitle={ICML},
  year={2024}
}

@inproceedings{chen2023polarnet,
  title={Polarnet: 3d point clouds for language-guided robotic manipulation},
  author={Chen, Shizhe and Garcia, Ricardo and Schmid, Cordelia and Laptev, Ivan},
  booktitle={CoRL},
  year={2023}
}

@inproceedings{guo2024phygrasp,
  title={PhyGrasp: generalizing robotic grasping with physics-informed large multimodal models},
  author={Guo, Dingkun and Xiang, Yuqi and Zhao, Shuqi and Zhu, Xinghao and Tomizuka, Masayoshi and Ding, Mingyu and Zhan, Wei},
  booktitle={IROS},
  year={2025}
}

@article{wu2026hug,
  title={Human Universal Grasping},
  author={Wu, Kevin Yuanbo and Zhou, Tianxing and Tu, Isaac and Yan, Billy and Guzey, Irmak and Fouhey, David and Shan, Dandan and Pinto, Lerrel},
  journal={arXiv preprint arXiv:2606.17054},
  year={2026}
}

@article{kolda2003optimization,
  title={Optimization by direct search: New perspectives on some classical and modern methods},
  author={Kolda, Tamara G and Lewis, Robert Michael and Torczon, Virginia},
  journal={SIAM review},
  year={2003}
}

@book{conn2009introduction,
  title={Introduction to derivative-free optimization},
  author={Conn, Andrew R and Scheinberg, Katya and Vicente, Luis N},
  year={2009},
  publisher={SIAM}
}

@article{rios2013derivative,
  title={Derivative-free optimization: a review of algorithms and comparison of software implementations},
  author={Rios, Luis Miguel and Sahinidis, Nikolaos V},
  journal={Journal of Global Optimization},
  year={2013}
}

@article{ghadimi2013stochastic,
  title={Stochastic first-and zeroth-order methods for nonconvex stochastic programming},
  author={Ghadimi, Saeed and Lan, Guanghui},
  journal={SIAM journal on optimization},
  year={2013}
}

@article{duchi2015optimal,
  title={Optimal rates for zero-order convex optimization: The power of two function evaluations},
  author={Duchi, John C and Jordan, Michael I and Wainwright, Martin J and Wibisono, Andre},
  journal={IEEE Transactions on Information Theory},
  year={2015},
}

@article{shahriari2015taking,
  title={Taking the human out of the loop: A review of Bayesian optimization},
  author={Shahriari, Bobak and Swersky, Kevin and Wang, Ziyu and Adams, Ryan P and De Freitas, Nando},
  journal={Proceedings of the IEEE},
  year={2015},
  publisher={IEEE}
}

@inproceedings{sun2022bbt,
  title={Black-box tuning for language-model-as-a-service},
  author={Sun, Tianxiang and Shao, Yunfan and Qian, Hong and Huang, Xuanjing and Qiu, Xipeng},
  booktitle={ICML},
  year={2022}
}

@inproceedings{xiao2020pca,
  title={Endowing deep 3d models with rotation invariance based on principal component analysis},
  author={Xiao, Zelin and Lin, Hongxin and Li, Renjie and Geng, Lishuai and Chao, Hongyang and Ding, Shengyong},
  booktitle={IEEE international conference on multimedia and expo},
  year={2020}
}

@inproceedings{sun2022bbtv2,
  title={Bbtv2: Towards a gradient-free future with large language models},
  author={Sun, Tianxiang and He, Zhengfu and Qian, Hong and Zhou, Yunhua and Huang, Xuan-Jing and Qiu, Xipeng},
  booktitle={EMNLP},
  year={2022}
}

@inproceedings{park2025zip,
  title={Zip: An efficient zeroth-order prompt tuning for black-box vision-language models},
  author={Park, Seonghwan and Jeong, Jaehyeon and Kim, Yongjun and Lee, Jaeho and Lee, Namhoon},
  booktitle={ICLR},
  year={2025}
}

@article{hansen2003reducing,
  title={Reducing the time complexity of the derandomized evolution strategy with covariance matrix adaptation (CMA-ES)},
  author={Hansen, Nikolaus and M{\"u}ller, Sibylle D and Koumoutsakos, Petros},
  journal={Evolutionary computation},
  year={2003}
}

@article{hansen2016cma,
  title={The CMA evolution strategy: A tutorial},
  author={Hansen, Nikolaus},
  journal={arXiv preprint arXiv:1604.00772},
  year={2016}
}

@inproceedings{oh2023blackvip,
  title={Blackvip: Black-box visual prompting for robust transfer learning},
  author={Oh, Changdae and Hwang, Hyeji and Lee, Hee-young and Lim, YongTaek and Jung, Geunyoung and Jung, Jiyoung and Choi, Hosik and Song, Kyungwoo},
  booktitle={CVPR},
  year={2023}
}

@inproceedings{malladi2023mezo,
  title={Fine-tuning language models with just forward passes},
  author={Malladi, Sadhika and Gao, Tianyu and Nichani, Eshaan and Damian, Alex and Lee, Jason D and Chen, Danqi and Arora, Sanjeev},
  booktitle={NeurIPS},
  year={2023}
}

@inproceedings{koh2020cbm,
  title={Concept bottleneck models},
  author={Koh, Pang Wei and Nguyen, Thao and Tang, Yew Siang and Mussmann, Stephen and Pierson, Emma and Kim, Been and Liang, Percy},
  booktitle={ICML},
  year={2020}
}

@inproceedings{park2025cbmnoise,
  title={An analysis of concept bottleneck models: Measuring, understanding, and mitigating the impact of noisy annotations},
  author={Park, Seonghwan and Mun, Jueun and Oh, Donghyun and Lee, Namhoon},
  booktitle={NeurIPS},
  year={2025}
}

@inproceedings{yang2022openood,
  title={Openood: Benchmarking generalized out-of-distribution detection},
  author={Yang, Jingkang and Wang, Pengyun and Zou, Dejian and Zhou, Zitang and Ding, Kunyuan and Peng, Wenxuan and Wang, Haoqi and Chen, Guangyao and Li, Bo and Sun, Yiyou and Du, Xuefeng and Zhou, Kaiyang and Zhang, Wayne and Hendrycks, Dan and Li, Yixuan and Liu, Ziwei},
  booktitle={NeurIPS Datasets and Benchmarks Track},
  year={2022}
}

@article{ren2021rmds,
  title={A simple fix to mahalanobis distance for improving near-ood detection},
  author={Ren, Jie and Fort, Stanislav and Liu, Jeremiah and Roy, Abhijit Guha and Padhy, Shreyas and Lakshminarayanan, Balaji},
  journal={arXiv preprint arXiv:2106.09022},
  year={2021}
}

@inproceedings{park2026fold,
  title={Exploiting Local Flatness for Efficient Out-of-Distribution Detection},
  author={Park, Seonghwan and Jung, Hyunji and Lee, Dongyeop and Lee, Namhoon},
  booktitle={ECCV},
  year={2026}
}
}
\clearpage
\appendix
\section{Additional Results}
\label{app:results}
This section presents additional experiments supporting \Cref{sec:dynamics}.
We first study optimizer behavior and query efficiency.
We then extend the augmentation-gap analysis to part segmentation and additional architectures.
Finally, we examine surrogate reliability and identity-prior safety.

\subsection{Optimizer and Query Budget}
\label{query_budget}

\begin{figure}[h]\centering
\includegraphics[width=0.95\linewidth]{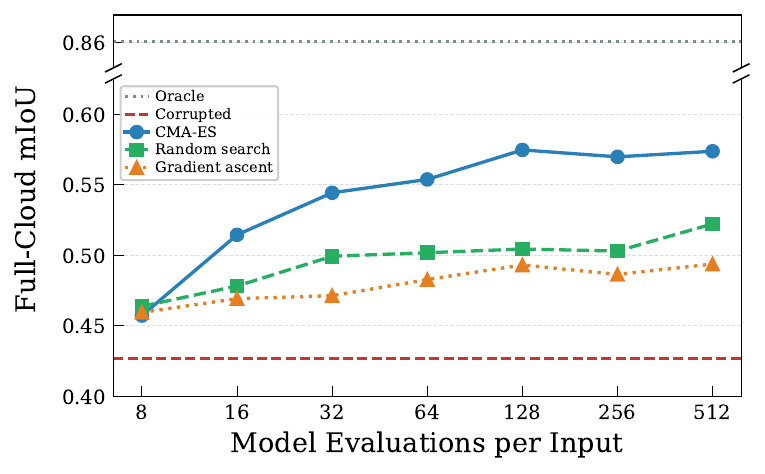}
\vspace{-0.75em}
\caption{Recovery versus optimization budget on a common model-evaluation axis. Gradient ascent plateaus well below global search, whereas CMA-ES is the most query-efficient.}
\label{fig:budget}
\end{figure}

\paragraph{Does local search catch up with budget?}
A natural question is whether the baselines are simply under-budgeted.
Figure~\ref{fig:budget} compares CMA-ES, random search, and white-box gradient ascent under a common model-evaluation budget.
Performance is reported as the best solution at each budget.
This sweep uses a smaller evaluation subset, so we focus on relative trends.
Even after $512$ evaluations, local search remains well below global search.
Gradient ascent plateaus at $0.494$ mIoU, whereas random search and CMA-ES reach $0.522$ and $0.574$ mIoU.
The three-restart variant of \Cref{tab:compare} reaches a similar plateau of $0.500$ mIoU, indicating that restarts do not change this behavior.
As shown in \Cref{tab:appbudget}, extending the budget to $8192$ evaluations leads to the same conclusion.
CMA-ES and random search saturate after roughly $128$ evaluations, whereas gradient ascent reaches only $0.528$ mIoU.
Both global methods exceed this value after only $32$ evaluations.
Increasing the budget does not rescue local optimization.

\begin{figure}[h]\centering
\includegraphics[width=\linewidth]{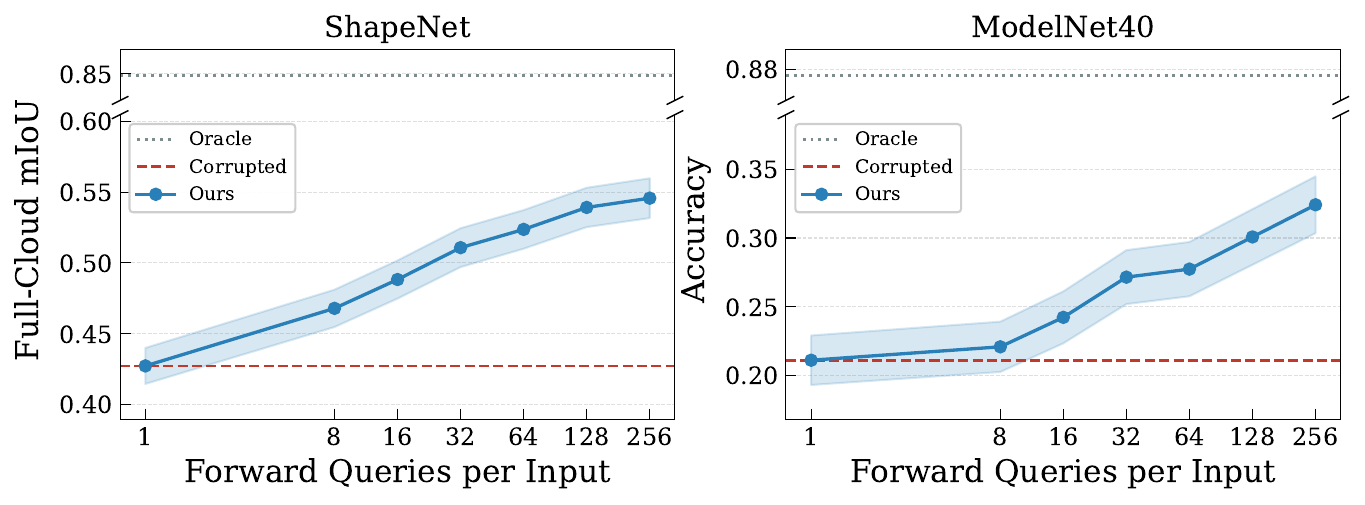}
\vspace{-0.75em}
\caption{Recovery versus query budget under random-axis $SO(3)$. Recovery shows diminishing returns and remains bounded by the oracle. Shaded bands denote standard error.}
\label{fig:qbudget}
\end{figure}

\paragraph{Query-budget curve.}
We then examine how recovery scales with the number of forward queries.
As shown in \Cref{fig:qbudget}, recovery increases monotonically with diminishing returns.
On part segmentation, performance improves from $0.43$ to $0.49$ with only $16$ queries and reaches $0.55$ with $256$ queries.
Classification follows the same trend, improving from $0.21$ to $0.32$.
Most of the achievable gain therefore arrives within a few dozen queries.
This supports the practical viability of recovery under limited query budgets.

\begin{table}[h]\centering\small
\caption{Budget-saturation study. Each row varies its own budget and is comparable only within the row. Increasing the budget provides little additional benefit across methods.}
\label{tab:appbudget}
\vspace{-0.75em}
\resizebox{0.95\linewidth}{!}{%
\begin{tabular}{llccc}
\toprule
\textbf{Method} & \textbf{Budgets} & \multicolumn{3}{c}{\textbf{Results at Each Budget}} \\
\midrule
TENT~\citep{tent}              & $1$ / $10$ / $50$     & 0.345 & 0.346 & 0.301 \\
MEMO~\citep{memo}              & $1$ / $10$ / $50$     & 0.344 & 0.357 & 0.337 \\
Gradient Ascent                & $32$ / $128$ / $8192$ & 0.509 & 0.515 & 0.528 \\
Random Search~\citep{bergstra2012random} & $32$ / $128$ / $8192$ & 0.541 & 0.575 & 0.567 \\
CMA-ES~\citep{hansen2001cmaes} & $32$ / $128$ / $8192$ & 0.552 & 0.602 & 0.577 \\
\bottomrule
\end{tabular}%
}
\end{table}

\subsection{Augmentation Gap}
\label{app:auggap}
\paragraph{Augmentation ablation on part segmentation.}
To verify that the augmentation-gap analysis of \Cref{subsec:auggap} extends beyond scene segmentation, we repeat the causal study on ShapeNetPart.
We train PointNet++ under three rotation-augmentation regimes and evaluate yaw, tilt, and full $SO(3)$ corruptions.
As shown in \Cref{tab:augablationseg}, the same trend holds for part segmentation.
Recovery is consistently larger when the training augmentation does not cover the test corruption.
Recovery becomes smaller once the corresponding augmentation is introduced.
Unlike classification, the clean cost of $SO(3)$ augmentation is modest.
Performance decreases from $0.842$ to $0.835$ mIoU.
On this task, the motivation for test-time recovery stems primarily from the frozen-model constraint rather than the clean-performance trade-off.

\begin{table}[h]\centering\small
\caption{Augmentation ablation on ShapeNetPart part segmentation. Each cell reports corrupted $\to$ recovered mIoU. Recovery decreases as augmentation coverage increases.}
\label{tab:augablationseg}
\vspace{-0.75em}
\resizebox{\linewidth}{!}{%
\begin{tabular}{ccccc}
\toprule
\textbf{Augmentation} & \textbf{Clean} & \textbf{Yaw} & \textbf{Tilt} & \textbf{$SO(3)$} \\
\midrule
None    & 0.842 & \cellcolor{red!44}$0.376{\to}0.540$ & \cellcolor{red!30}$0.520{\to}0.621$ & \cellcolor{red!33}$0.469{\to}0.586$ \\
Yaw     & 0.838 & \cellcolor{red!6}$0.840{\to}0.842$ & \cellcolor{red!27}$0.539{\to}0.628$ & \cellcolor{red!37}$0.547{\to}0.682$ \\
$SO(3)$ & 0.835 & \cellcolor{red!6}$0.820{\to}0.828$ & \cellcolor{red!6}$0.822{\to}0.827$ & \cellcolor{red!6}$0.819{\to}0.827$ \\
\bottomrule
\end{tabular}%
}
\end{table}

\paragraph{Cross-architecture sensitivity.}
We next examine whether transformation sensitivity generalizes across architectures.
Figure~\ref{fig:pitch_b} shows the tilt-sensitivity curve of the scene segmentation model.
Increasing tilt monotonically reduces mIoU from $0.627$ when upright to $0.088$ at $60^\circ$.
By contrast, yaw rotations of the same magnitude leave performance unchanged.
Table~\ref{tab:apptilt} extends the comparison across architectures.
All three backbones exhibit similar tilt sensitivity while remaining robust to yaw, including vanilla PointNet~\citep{qi2017pointnet}.
The two scalable models~\citep{qi2017pointnetpp, qian2022pointnext} are also robust within their trained scale range.
These results indicate that training augmentation, rather than architecture, determines robustness.

\begin{table}[h]\centering\small
\caption{Accuracy drops under tilt, scale, and color across architectures. All backbones remain robust to augmented transformations but degrade substantially under the unseen tilt.}
\label{tab:apptilt}
\vspace{-0.75em}
\resizebox{0.75\linewidth}{!}{%
\begin{tabular}{lccc}
\toprule
\textbf{Drop Under} & \textbf{Tilt} & \textbf{Large Scale} & \textbf{Color} \\
\midrule
PointNet~\citep{qi2017pointnet}      & \cellcolor{red!16}0.48 & -- & -- \\
PointNet++~\citep{qi2017pointnetpp}  & \cellcolor{red!16}0.42 & 0.20 & 0.09 \\
PointNeXt-S~\citep{qian2022pointnext} & \cellcolor{red!16}0.53 & 0.20 & 0.08 \\
\bottomrule
\end{tabular}%
}
\end{table}

\paragraph{Cross-architecture recovery.}
We then test whether recovery also generalizes across architectures.
We repeat the $SO(3)$ classification experiment of \Cref{tab:canon} on three additional non-equivariant backbones spanning MLP, convolutional, and transformer families~\citep{qi2017pointnetpp, qian2022pointnext, deng2022pointvector, pang2022pointmae}.
Each backbone is frozen and accessed only through forward queries.
As shown in \Cref{tab:clsarch}, all models suffer severe degradation under $SO(3)$ rotations.
\cando consistently recovers a substantial portion of the oracle gap.
The weaker recovery therefore stems from the single-softmax objective rather than the architecture.
Point-MAE supports this conclusion by showing the same behavior as the other backbones.
The confidence-accuracy correlation also remains positive across backbones, indicating that the recovery condition generalizes across architectures.

\begin{table}[h]\centering\small
\caption{Classification recovery under $SO(3)$ rotations across architecture families. \cando consistently recovers a substantial portion of the oracle gap, and confidence-accuracy correlation remains positive across backbones.}
\label{tab:clsarch}
\vspace{-0.75em}
\resizebox{\linewidth}{!}{%
\begin{tabular}{lcccccc}
\toprule
\textbf{Backbone} & \textbf{Family} & \textbf{Clean} & \textbf{$SO(3)$} & \textbf{$+$\cando} & \textbf{\% Gap} & \textbf{Corr} \\
\midrule
PointNet++~\citep{qi2017pointnetpp}    & MLP         & 0.92 & 0.13 & 0.34 & \cellcolor{green!9}29 & 0.31 \\
PointNeXt-S~\citep{qian2022pointnext}  & Conv        & 0.93 & 0.16 & \textbf{0.48} & \cellcolor{green!14}\textbf{42} & 0.52 \\
PointVector-S~\citep{deng2022pointvector} & Conv     & 0.93 & 0.15 & 0.47 & \cellcolor{green!14}40 & 0.54 \\
Point-MAE~\citep{pang2022pointmae}     & Transformer & 0.93 & 0.15 & 0.41 & \cellcolor{green!11}34 & 0.26 \\
\bottomrule
\end{tabular}%
}
\end{table}

\subsection{Surrogate Reliability}
\label{app:surrogate boundary}
\paragraph{Full corruption benchmark.}
The main text reports per-corruption ModelNet-C results for classification in \Cref{tab:appmc}.
We further evaluate the segmentation model on all seven ShapeNet-C corruptions in \Cref{tab:appsc}.
The dense per-point objective does not degrade performance at full evaluation.
Every change is non-negative and at most $+0.004$ mIoU.
On shared corruptions, the segmentation objective keeps the identity transform for $86$-$92\%$ of inputs.
Classification keeps the identity for only $50$-$80\%$.
The qualitative pattern is stable: dense confidence is less susceptible to harmful overconfidence than the single-softmax classification signal.

\begin{table}[h]\centering\small
\caption{ShapeNet-C recovery with the part-segmentation model. The dense per-point confidence surrogate remains reliable across all corruption families, with every change between $+0.001$ and $+0.004$ mIoU.}
\label{tab:appsc}
\vspace{-0.75em}
\resizebox{\linewidth}{!}{%
\begin{tabular}{llccc}
\toprule
\textbf{Corruption} & \textbf{Family} & \textbf{Source} & \textbf{$+$\cando} & \textbf{$\Delta$} \\
\midrule
Rotate          & Transform & 0.805 & 0.806 & \cellcolor{green!8}$+0.001$ \\
Scale           & Transform & 0.839 & 0.840 & \cellcolor{green!8}$+0.001$ \\
Jitter          & Noise     & 0.791 & 0.795 & \cellcolor{green!8}$+0.004$ \\
Add (Global)    & Noise     & 0.813 & 0.815 & \cellcolor{green!8}$+0.002$ \\
Add (Local)     & Noise     & 0.837 & 0.839 & \cellcolor{green!8}$+0.002$ \\
Dropout (Global)& Density   & 0.850 & 0.852 & \cellcolor{green!8}$+0.001$ \\
Dropout (Local) & Density   & 0.821 & 0.822 & \cellcolor{green!8}$+0.001$ \\
\midrule
Mean Over Seven & --        & 0.822 & 0.824 & \cellcolor{green!8}$+0.002$ \\
\bottomrule
\end{tabular}%
}
\end{table}

\paragraph{Calibration and logit access.}
Because \cando optimizes model confidence, its performance could depend on score calibration or on whether the API exposes probabilities rather than logits.
We evaluate both factors in \Cref{tab:appcalib}.
We re-run $SO(3)$ recovery on ModelNet40 using temperature-scaled confidence, maximum log-probability, and negative entropy.
Recovery remains stable from sharpening to mild softening.
It ranges from $0.297$ to $0.309$ for $T\in[0.5,2]$ and drops only under extreme softening to $0.277$ at $T{=}4$.
The log-probability objective matches probability access at $0.309$ because selecting the best rotation is invariant to monotone score transformations.
Negative entropy reaches $0.307$, and it also covers the divergence from the uniform distribution $\bm{u}$, since $\mathrm{KL}(\bm{p}\,\|\,\bm{u}) = \log C - \mathcal{H}(\bm{p})$ for $C$ classes and therefore ranks candidates identically.
These results indicate that \cando requires only an ordinal confidence signal rather than calibrated probabilities.

\begin{table}[h]\centering\small
\caption{Calibration study of $SO(3)$ recovery on ModelNet40. Recovery remains stable across temperature scaling and is identical under logit and probability access.}
\label{tab:appcalib}
\vspace{-0.75em}
\resizebox{\linewidth}{!}{%
\begin{tabular}{lccccccc}
\toprule
\textbf{Objective} & \textbf{Prob.} & \textbf{Temp.\,$0.5$} & \textbf{Temp.\,$2$} & \textbf{Temp.\,$4$} & \textbf{Log-prob.} & \textbf{Neg.\,Entropy} \\
\midrule
Recovered & $0.309$ & $0.297$ & $0.309$ & $0.277$ & $0.309$ & $0.307$ \\
\bottomrule
\end{tabular}%
}
\end{table}

\begin{figure*}[h]\centering
\includegraphics[width=0.85\linewidth]{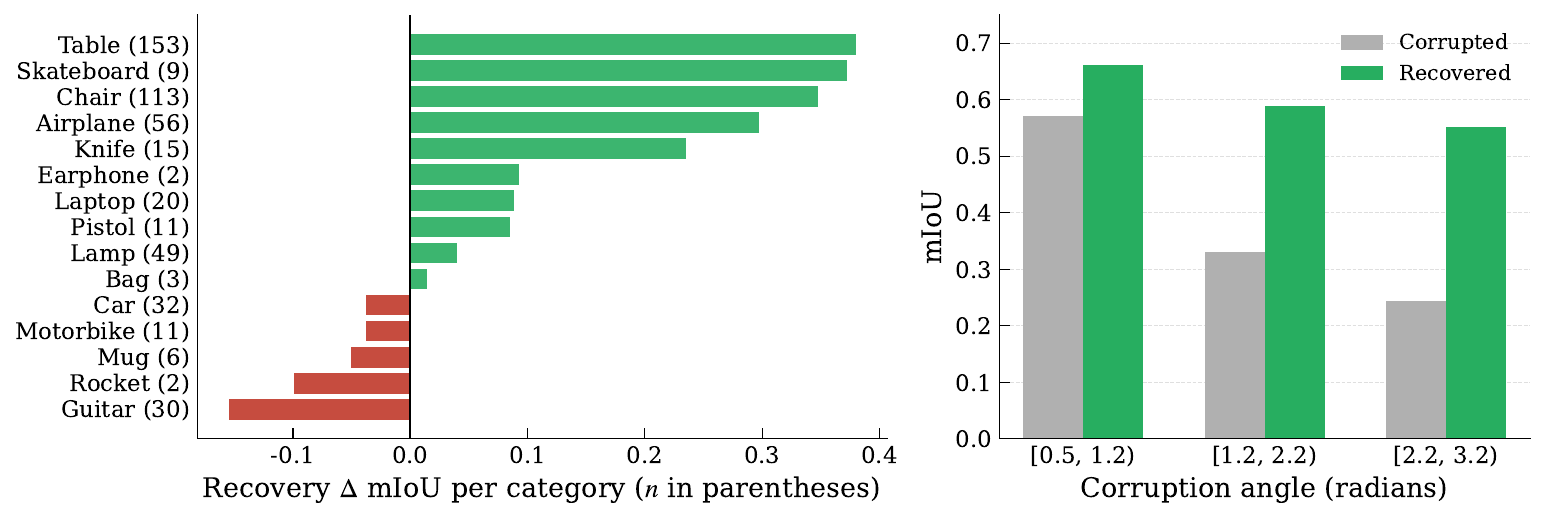}
\vspace{-0.75em}
\caption{Per-category analysis of $SO(3)$ recovery. (Left) Categories with distinctive canonical frames exhibit the largest gains, whereas rotationally ambiguous categories show limited or negative recovery. (Right) Recovery increases with corruption magnitude, indicating that residual failures arise primarily from geometric ambiguity rather than optimization.}
\label{fig:percat}
\end{figure*}

\paragraph{Per-category and failure analysis.}
\Cref{fig:percat} breaks the $SO(3)$ recovery study down by category.
Gains concentrate on categories with distinctive canonical frames.
Tables improve by $0.38$ mIoU, skateboards by $0.37$, chairs by $0.35$, and airplanes by $0.30$.
Losses instead concentrate on rotationally ambiguous shapes.
Guitars lose $0.16$, rockets $0.10$, mugs $0.05$, and cars $0.04$.
For these categories, multiple orientations explain the geometry equally well.
The confidence surrogate can therefore favor an incorrect pose.
Overall, $63\%$ of shapes improve by more than $0.01$ mIoU and $26\%$ degrade by the same margin.
The mean change is $+0.23$.
Recovery also increases with corruption magnitude.
Inputs below $1.2$ radians improve from $0.571$ to $0.662$ mIoU, whereas inputs above $2.2$ radians improve from $0.244$ to $0.551$ mIoU.
These results indicate that the remaining failures arise primarily from geometric ambiguity rather than optimization.

\subsection{Identity Prior and Safety}
\label{app:identity prior}

\paragraph{Do-no-harm on clean inputs.}
Beyond the average trade-off in \Cref{fig:recangle}, we measure how often recovery harms already well-aligned inputs.
As shown in \Cref{tab:dnh}, the identity prior ($\lambda{=}0.08$) limits degradation beyond $1\%$ to only $6\%$ of clean inputs.
The mean change is $-0.005$ mIoU for part segmentation.
Without the prior, $41\%$ of inputs degrade by more than $1\%$ and the mean change falls to $-0.143$.
The identity prior therefore suppresses unnecessary transformations on clean inputs while preserving recovery, making it essential for safe deployment.

\begin{table}[h]\centering\small
\caption{Tail risk on clean inputs for part segmentation and classification: mean accuracy change and the fraction of inputs degraded by more than $1$, $5$, and $10\%$, with and without the identity prior.}
\label{tab:dnh}
\vspace{-0.75em}
\resizebox{\linewidth}{!}{%
\begin{tabular}{llcccc}
\toprule
\textbf{Task} & \textbf{Prior} & \textbf{Mean $\Delta$} & \textbf{$>1\%$ harmed} & \textbf{$>5\%$ harmed} & \textbf{$>10\%$ harmed} \\
\midrule
\rowcolor{green!6}
Part Seg. & $\lambda{=}0.08$ & $-0.005$ & $6.2\%$  & $3.2\%$  & $2.6\%$ \\
\rowcolor{red!6}
Part Seg. & $\lambda{=}0$    & $-0.143$ & $41.0\%$ & $31.6\%$ & $29.1\%$ \\
\rowcolor{green!6}
Classif. & $\lambda{=}0.08$ & $-0.053$ & $6.0\%$  & $6.0\%$  & $6.0\%$ \\
\rowcolor{red!6}
Classif. & $\lambda{=}0$    & $-0.231$ & $23.4\%$ & $23.4\%$ & $23.4\%$ \\
\bottomrule
\end{tabular}%
}
\end{table}

\paragraph{Identity-prior ablation.}
\Cref{tab:applam} evaluates different values of the identity-prior weight $\lambda$.
Larger values preserve the identity transformation more often while maintaining rotation recovery.
They also reduce degradation on non-pose corruptions.
Outlier addition remains the most difficult corruption because the confidence surrogate becomes overconfident without improving correctness.
Smaller values increase harm on jitter and outlier addition without improving rotation recovery.
They are therefore beneficial only when inputs are known to contain large pose errors.

\begin{table}[h]\centering\small
\caption{Identity-prior ablation: change in accuracy versus the prior weight $\lambda$ on ModelNet-C. Larger $\lambda$ preserves the rotation gain while reducing harm on non-pose corruptions.}
\label{tab:applam}
\vspace{-0.75em}
\resizebox{0.8\linewidth}{!}{%
\begin{tabular}{lcccc}
\toprule
\textbf{$\lambda$} & \textbf{Rotate} & \textbf{Scale} & \textbf{Jitter} & \textbf{Add (Global)} \\
\midrule
0.2 & \cellcolor{green!12}$+0.086$ & \cellcolor{red!8}$-0.031$ & \cellcolor{red!8}$-0.086$ & \cellcolor{red!12}$-0.148$ \\
0.4 & \cellcolor{green!12}$+0.047$ & \cellcolor{red!4}$-0.016$ & \cellcolor{red!8}$-0.055$ & \cellcolor{red!12}$-0.102$ \\
0.7 & \cellcolor{green!12}$+0.031$ & \cellcolor{red!4}$-0.016$ & \cellcolor{red!8}$-0.047$ & \cellcolor{red!8}$-0.062$ \\
\bottomrule
\end{tabular}%
}
\end{table}

\paragraph{Confidence-margin gate.}
\Cref{tab:appgate} evaluates a confidence-margin gate that accepts a recovered rotation only if its confidence exceeds the identity by a margin $\tau$.
The gate preserves rotation recovery while reducing degradation on jitter and dropout.
It has little effect on global outlier addition because the surrogate remains confidently incorrect.
This limitation therefore reflects the confidence surrogate rather than the acceptance threshold.
Gating adaptation by predictive uncertainty parallels concept bottleneck models~\citep{koh2020cbm}, where correcting only the most uncertain concepts mitigates annotation noise~\citep{park2025cbmnoise}.
Lightweight post-hoc OOD detectors~\citep{yang2022openood,ren2021rmds,park2026fold} could further flag non-pose corruptions before search.

\begin{table}[h]\centering\small
\caption{Confidence-margin gate on ModelNet-C: change in accuracy versus the acceptance margin $\tau$, with $\tau{=}0$ the ungated method. Larger $\tau$ reduces harm on corruptions with misaligned confidence while preserving the rotation gain, but cannot fix outlier addition.}
\label{tab:appgate}
\vspace{-0.75em}
\resizebox{\linewidth}{!}{%
\begin{tabular}{llcccc}
\toprule
\textbf{Corruption} & \textbf{Family} & \textbf{$\tau{=}0$} & \textbf{$\tau{=}0.05$} & \textbf{$\tau{=}0.1$} & \textbf{$\tau{=}0.2$} \\
\midrule
Rotate          & Transform & \cellcolor{green!12}$+0.060$ & \cellcolor{green!12}$+0.058$ & \cellcolor{green!12}$+0.067$ & \cellcolor{green!12}$+0.054$ \\
Jitter          & Noise     & \cellcolor{red!8}$-0.067$ & \cellcolor{red!8}$-0.058$ & \cellcolor{red!8}$-0.048$ & \cellcolor{red!8}$-0.067$ \\
Add (Global)    & Noise     & \cellcolor{red!8}$-0.073$ & \cellcolor{red!8}$-0.065$ & \cellcolor{red!8}$-0.067$ & \cellcolor{red!8}$-0.060$ \\
Dropout (Global)& Density   & \cellcolor{red!4}$-0.017$ & \cellcolor{red!4}$-0.017$ & \cellcolor{green!4}$-0.002$ & \cellcolor{green!4}$-0.002$ \\
\bottomrule
\end{tabular}%
}
\end{table}

\paragraph{Mixed deployment stream.}
\Cref{tab:appmixed} evaluates a single acceptance policy on a mixed deployment stream containing canonical, mildly rotated, and severely rotated inputs.
Increasing the confidence threshold leaves more canonical inputs untouched but gradually reduces recovery on rotated ones.
Part segmentation admits an operating point that preserves canonical inputs while maintaining a positive net gain.
Classification instead exhibits an unavoidable trade-off because confidently incorrect rotations can satisfy any practical threshold.
The reported results therefore characterize the deployment range of a single fixed policy for each surrogate.

\begin{table}[h]\centering\small
\caption{Mixed-stream single-policy sweep, averaged over three seeds. Part segmentation admits a do-no-harm operating point with positive net gain, whereas classification trades canonical harm for the rotated gain at every threshold.}
\label{tab:appmixed}
\vspace{-0.75em}
\resizebox{\linewidth}{!}{%
\begin{tabular}{llccccc}
\toprule
\textbf{Task} & \textbf{$\tau$} & \textbf{Net Gain} & \textbf{Clean $\Delta$} & \textbf{Mild $\Delta$} & \textbf{Severe $\Delta$} & \textbf{Identity Kept} \\
\midrule
\multirow{4}{*}{Part Seg.} & $0.00$ & \cellcolor{green!12}$+0.036$ & \cellcolor{red!8}$-0.002$ & \cellcolor{green!16}$+0.081$ & \cellcolor{green!14}$+0.067$ & $70\%$ \\
                           & $0.05$ & \cellcolor{green!12}$+0.030$ & \cellcolor{red!8}$-0.004$ & \cellcolor{green!14}$+0.066$ & \cellcolor{green!14}$+0.064$ & $98\%$ \\
                           & $0.10$ & \cellcolor{green!10}$+0.021$ & \cellcolor{red!8}$-0.003$ & \cellcolor{green!12}$+0.039$ & \cellcolor{green!14}$+0.051$ & $99\%$ \\
                           & $0.20$ & \cellcolor{green!8}$+0.003$ & $+0.000$ & \cellcolor{green!8}$+0.001$ & \cellcolor{green!10}$+0.013$ & $100\%$ \\
\midrule
\multirow{4}{*}{Classif.}  & $0.00$ & \cellcolor{green!10}$+0.028$ & \cellcolor{red!14}$-0.051$ & \cellcolor{green!18}$+0.156$ & \cellcolor{green!14}$+0.060$ & $70\%$ \\
                           & $0.05$ & \cellcolor{green!10}$+0.029$ & \cellcolor{red!12}$-0.049$ & \cellcolor{green!18}$+0.154$ & \cellcolor{green!14}$+0.060$ & $83\%$ \\
                           & $0.10$ & \cellcolor{green!12}$+0.031$ & \cellcolor{red!12}$-0.040$ & \cellcolor{green!18}$+0.146$ & \cellcolor{green!14}$+0.058$ & $87\%$ \\
                           & $0.20$ & \cellcolor{green!12}$+0.033$ & \cellcolor{red!10}$-0.025$ & \cellcolor{green!18}$+0.132$ & \cellcolor{green!14}$+0.052$ & $91\%$ \\
\bottomrule
\end{tabular}%
}
\end{table}



\section{Experimental Details}
\label{app:setup}

This section provides additional details on the experimental framework, including baselines, optimizer comparisons, implementation details, and dataset preprocessing.

\subsection{Baselines}
\label{app:baselines}

We compare \cando with gradient-free methods compatible with black-box access and gradient-based methods that serve as white-box references.
This section expands the setup in \Cref{subsec:setup} by specifying the rule used by each baseline.

\paragraph{Principal-axis canonicalization~\citep{xiao2020pca}.}
This baseline aligns the input to its principal axes.
It forms the centered coordinate covariance and eigendecomposes it,
\begin{equation}
\Cb = \frac{1}{N}\sum_{i=1}^{N}(\bp_i - \bar{\bp})(\bp_i - \bar{\bp})^{\top} = \Vb\bm{\Lambda}\Vb^{\top},
\end{equation}
then canonicalizes each point as $\hat{\bp}_i = \Vb^{\top}(\bp_i - \bar{\bp})$.
Because each eigenvector is defined only up to sign, the canonical frame is ambiguous under the $2^{3}$ sign flips of the columns of $\Vb$.
The confidence-disambiguated variant resolves this ambiguity by selecting the flip that maximizes the model confidence $J$.

\paragraph{Normalization-statistic adaptation~\citep{mirza2022dua}.}
DUA keeps all weights fixed and updates only batch-normalization statistics.
For each normalization layer, it blends the stored running statistics $(\bmu, \bm{\sigma}^{2})$ with the test-input statistics $(\bmu_{\Pb}, \bm{\sigma}_{\Pb}^{2})$,
\begin{equation}
\bmu \leftarrow (1-\rho)\,\bmu + \rho\,\bmu_{\Pb}, \qquad
\bm{\sigma}^{2} \leftarrow (1-\rho)\,\bm{\sigma}^{2} + \rho\,\bm{\sigma}_{\Pb}^{2},
\end{equation}
then recomputes the prediction with the adapted statistics.
We use $16$ augmented copies of the input and apply three updates with momentum $\rho{=}0.1$.

\paragraph{External-prior canonicalization.}
This baseline keeps the pose search of \cando but replaces the scoring signal.
It selects
\begin{equation}
\hat{\btheta} = \argmax_{\btheta} J'(\btheta),
\end{equation}
where $J'$ is the confidence of a separate frozen classifier $g'$ rather than the task model.
It then returns the task prediction $g(\Tb_{\hat{\btheta}}(\Pb))$ at the selected pose.
This isolates the effect of the scoring model at the cost of an auxiliary network.
We use a frozen ModelNet40 PointNet++ classifier as $g'$ and therefore report this baseline only on ShapeNetPart, since an object classifier provides no meaningful pose prior for S3DIS scene blocks and on ModelNet40 it would coincide with the task model.

\paragraph{TENT~\citep{tent}.}
TENT takes gradient-descent steps on the Shannon entropy of the prediction $\bm{p} = g(\Pb)$,
\begin{equation}
\begin{aligned}
\min_{\{\bm{\gamma},\bm{\beta}\}} \quad
& \mathcal{H}(\bm{p}) \\
\text{s.t.}\quad
& \mathcal{H}(\bm{p})=-\sum_{c}p_c\log p_c .
\end{aligned}
\end{equation}
updating only the affine normalization parameters $\{\bm{\gamma}, \bm{\beta}\}$.
All other weights remain fixed.
We use SGD steps with momentum $0.9$ and learning rate $5{\times}10^{-4}$ for segmentation and $10^{-3}$ for classification.

\paragraph{MEMO~\citep{memo}.}
MEMO minimizes the marginal entropy of predictions over augmentations $\{a_{b}\}_{b=1}^{B}$,
\begin{equation}
\begin{aligned}
\min \quad
& \mathcal{H}(\bar{\bm p}) \\
\text{s.t.}\quad
& \bar{\bm p}
=
\frac{1}{B}
\sum_{b=1}^{B}
g(a_b(\Pb)).
\end{aligned}
\end{equation}
and updates model parameters to encourage confident and augmentation-consistent outputs.
We use $B{=}8$ augmentations for segmentation and $B{=}16$ for classification.
The SGD settings match TENT.

\paragraph{Gradient-based canonicalization.}
This white-box reference applies \cando's pose search with exact gradients.
It ascends the confidence objective over the input rotation,
\begin{equation}
\btheta_{t+1} = \btheta_{t} + \eta\,\nabla_{\btheta} J(\btheta_{t}),
\end{equation}
where gradients are backpropagated through the model.
We run $20$ ascent steps with step size $0.2$ from three restarts under a query budget matched to \cando.

\paragraph{Oracle.}
The oracle provides a supervised upper bound.
For each input, it selects
\begin{equation}
\btheta^{\star} = \argmax_{\btheta} \mathcal{M}\!\left(g(\Tb_{\btheta}(\Pb)),\, y\right),
\end{equation}
where $\mathcal{M}$ is the true task metric against the ground-truth label $y$.
We use mIoU for segmentation and accuracy for classification.
No label-free method can exceed this oracle.

\subsection{Optimization Baselines}
\label{app:optimizers}

The optimizer study in \Cref{subsec:optimizer} compares CMA-ES with local gradient estimators and alternative global searches.
All methods use only forward queries of the confidence objective $J(\btheta)$.
Each baseline uses three random restarts under a query budget matched to \cando.
Zeroth-order methods have also been made query- and memory-efficient for tuning large black-box models~\citep{oh2023blackvip,malladi2023mezo,park2025zip}; such estimators may matter for transformation families richer than the three-dimensional rotations considered here.

\paragraph{Coordinate finite differences~\citep{conn2009introduction}.}
This estimator probes each axis independently with step size $\varepsilon$.
The $i$-th component is estimated by the central difference
\begin{equation}
\hat{g}_{i} = \frac{J(\btheta + \varepsilon\be_{i}) - J(\btheta - \varepsilon\be_{i})}{2\varepsilon},
\end{equation}
and the optimizer ascends along the assembled gradient $\hat{\bg}$.
We set $\varepsilon{=}0.1$.

\paragraph{SPSA~\citep{spall1992spsa}.}
SPSA estimates the full gradient using a single random Rademacher perturbation $\bDelta$ of magnitude $c$.
The estimator is
\begin{equation}
\hat{\bg}
=
\frac{J(\btheta+c\bDelta)-J(\btheta-c\bDelta)}
{2c}\,
\bDelta^{-1},
\end{equation}
where $\bDelta^{-1}$ denotes the componentwise inverse.
We set $c{=}0.1$.

\paragraph{Natural evolution strategies~\citep{wierstra2014nes}.}
NES smooths $J$ with a Gaussian search distribution of scale $\sigma$.
It estimates an antithetic search gradient from $m$ symmetric sample pairs:
\begin{equation}
\hat{\bg}
=
\frac{1}{\sigma m}
\sum_{k=1}^{m}
\Bigl[
J(\btheta+\sigma\bepsilon_k)
-
J(\btheta-\sigma\bepsilon_k)
\Bigr]
\bepsilon_k ,
\end{equation}
where $\bepsilon_k \sim \mathcal{N}(\bm0,\Ib)$ are i.i.d. Gaussian perturbations.
The symmetric pairs reduce variance by canceling odd components around $\btheta$.
We use $\sigma{=}0.15$ and $m{=}4$ pairs.

\paragraph{Score-function estimator~\citep{williams1992reinforce}.}
The score-function estimator uses the same Gaussian smoothing with one-sided samples and a running baseline $b$:
\begin{equation}
\hat{\bg} = \frac{1}{\sigma m}\sum_{k=1}^{m} \bigl[J(\btheta + \sigma\bepsilon_{k}) - b\bigr]\,\bepsilon_{k}.
\end{equation}
This follows the REINFORCE log-derivative estimator.
The running mean score $b$ reduces variance.
We use $\sigma{=}0.15$ and $m{=}8$ samples.

\paragraph{Random search~\citep{bergstra2012random}.}
Random search samples a fixed pool of $M$ candidate rotations independently from a search prior $\pi$ and returns the highest-scoring candidate:
\begin{equation}
\hat{\btheta}
=
\argmax_k
J(\btheta_k),
\end{equation}
where $\btheta_k\!\sim\!\pi$ independently for all $k$.
We use $M\in\{96,128,256\}$ depending on the experiment.

\paragraph{Bayesian optimization~\citep{shahriari2015taking}.}
Bayesian optimization fits a Gaussian-process surrogate to observed pose-score pairs $\mathcal{D}_{t} = \{(\btheta_{j}, J(\btheta_{j}))\}$.
It selects each query by maximizing an acquisition function $\alpha$:
\begin{equation}
\btheta_{t+1} = \argmax_{\btheta} \alpha(\btheta; \mathcal{D}_{t}).
\end{equation}
This balances exploration and exploitation.

\subsection{Implementation Details}
\label{app:impl}

For every input, the front-end applies the rotation transform of Eq.~\eqref{eq:rodrigues}. Candidate transformations are scored by the confidence surrogate of Eq.~\eqref{eq:conf} and selected using the identity prior of Eq.~\eqref{eq:prior}. Full $SO(3)$ search uses CMA-ES with two restarts: a population of $12$, $\sigma_0{=}1.0$, and $300$ queries for PointNet++ part segmentation; a population of $10$, $\sigma_0{=}1.2$, and $300$ queries for PointNeXt-S part segmentation and the optimizer study; and a population of $10$ ($12$ in \Cref{tab:compare}), $\sigma_0{=}1.2$, and $240$ queries for classification. Azimuth recovery scans $16$ angles and refines the best one with $15$ SPSA steps. S3DIS tilt recovery evaluates a deterministic $13{\times}13$ grid over the two tilt parameters ($169$ queries) in \Cref{tab:canon} and uses CMA-ES with a population of $8$, $\sigma_0{=}0.5$, and $160$ queries in \Cref{tab:compare}. The identity-prior weight $\lambda$ is selected once on a small labeled validation split and fixed thereafter. A per-input random seed initializes every estimator. All experiments are run on a single RTX A6000.

\subsection{Datasets and Preprocessing}
\label{app:data}

\paragraph{ShapeNetPart~\citep{yi2016shapenetpart}.}
Each shape is normalized to the unit sphere and sampled as a point cloud with surface normals.
Coordinates and normals are used as input.
Predictions on transformed points are propagated to the original cloud by three-nearest-neighbor voting.
We report category-masked instance mIoU following the standard evaluation protocol.

\paragraph{S3DIS~\citep{armeni2016s3dis}.}
We evaluate Area~5 using the standard block-based preprocessing with $4096$-point nine-channel inputs.
Tilt corruption and recovery rotate the input coordinates and recompute all geometric channels from the transformed geometry.
Color channels remain unchanged.

\paragraph{ModelNet40~\citep{wu2015modelnet}.}
We use the standard resampled benchmark with $1024$ points per shape.
Each shape is normalized to the unit sphere and represented only by point coordinates.

\paragraph{Corruptions.}
Pose corruptions sample a uniform azimuth, a random-axis $SO(3)$ rotation, or a horizontal-axis tilt independently for each input.
The $SO(3)$ rotation angle is sampled uniformly above a small minimum rather than from the Haar measure.
Experiments under Haar-uniform rotations lead to the same conclusions.
The non-pose study of \Cref{subsec:gameable} uses the official PointCloud-C generators~\citep{pointcloudc} with seven corruption types at five severity levels.

\section{Limitations and Future Work}
\label{app:limitations}

\paragraph{Limitations.}
The method targets rotation-induced shifts and does not address noise or density corruptions when confidence is misaligned with task performance.
Classification recovers less than segmentation because a single softmax provides a weaker optimization signal than dense per-point confidence.
The method also assumes a low-dimensional input transformation, probability or logit access, and a moderate forward-query budget.
Finally, our evaluation considers synthetic pose shifts, leaving validation on real sensor tilt and outdoor LiDAR data to future work.

\paragraph{Future Work.}
A natural next step is to extend the transformation family beyond rotations to richer parameterizations such as affine transformations, or to infer the transformation family automatically rather than specifying it a priori. Another promising direction is to combine forward-query pose recovery with external generative or foundation-model priors, extending the framework to distribution shifts that cannot be resolved by a confidence-only objective.

\end{document}